%% file: main.tex
\documentclass[runningheads]{llncs}

\usepackage{eccv}

\newcommand{\authcount}[1]{}
\usepackage{eccvabbrv}

\usepackage{graphicx}
\usepackage{booktabs}
\usepackage{multirow}
\usepackage{makecell}
\usepackage[table]{xcolor}

\usepackage[accsupp]{axessibility}  

\input{math_commands.tex}

\input{tex/abbrev}

\usepackage{booktabs}
\usepackage{array}
\usepackage{adjustbox}

\usepackage{hyperref}

\usepackage{orcidlink}

\begin{document}

\title{Task Alignment: A Simple Proxy for Practical Model Merging Across Diverse Vision Tasks}

\titlerunning{Task Alignment}

\author{Pau de Jorge\and
C\'esar Roberto de Souza\and
Björn Michele\and
Mert Bülent Sarıyıldız\and
Philippe Weinzaepfel\and
Florent Perronnin\and
Diane Larlus\and
Yannis Kalantidis}

\authorrunning{P. de Jorge et al.}

\institute{
NAVER LABS Europe  
}
\maketitle

\input{tex/00_abstract}
\input{tex/01_intro}
\input{tex/02_related_work}
\input{tex/03_TAS}

\input{tex/05_results}

\input{tex/06_analysis}
\input{tex/07_conclusion}



%
%
\bibliographystyle{splncs04}
\bibliography{main}

\clearpage

\input{tex/08_appendix}

\end{document}

%% file: math_commands.tex
\usepackage{amsmath,amsfonts,bm}

\def\eqref#1{equation~\ref{#1}}

\def\1{\bm{1}}

\def\vs{{\bm{s}}}

\DeclareMathAlphabet{\mathsfit}{\encodingdefault}{\sfdefault}{m}{sl}
\SetMathAlphabet{\mathsfit}{bold}{\encodingdefault}{\sfdefault}{bx}{n}

\DeclareMathOperator*{\argmin}{arg\,min}

%% file: tex/abbrev.tex
\usepackage{xspace}
\usepackage[dvipsnames]{xcolor}
\usepackage{mfirstuc}
\usepackage{wrapfig}

\usepackage{pgfplots}
\pgfplotsset{compat=1.18}

\usepackage{enumitem}
\setitemize{noitemsep,topsep=0pt,parsep=0pt,partopsep=0pt}

\usepackage{titletoc}

\newcommand{\talign}{TAP\xspace}

\newcommand{\taskalign}{task alignment proxy\xspace}
\newcommand{\taskalignloss}{task alignment loss\xspace}
\newcommand{\Taskalign}{Task alignment proxy\xspace}
\newcommand{\TaskAlign}{Task Alignment Proxy\xspace}

\newcommand{\HS}{Eval\xspace}

\newcommand{\adamerge}{AdaMerging\xspace}
\newcommand{\tadamerge}{AdaMerging\textsubscript{+\talign}\xspace}

\newcommand{\nyudNS}{NYUd}
\newcommand{\bedlamNS}{Bedlam}

\newcommand{\ade}{ADE20k\xspace}
\newcommand{\nyud}{NYUd\xspace}
\newcommand{\bedlam}{Bedlam\xspace}
\newcommand{\mapfree}{MapFree\xspace}

\newcommand{\Finetuning}{Fine-tuning\xspace}
\newcommand{\finetuning}{fine-tuning\xspace}

\definecolor{deeppink}{RGB}{255,20,147} 

\makeatletter
\newcommand{\CapitalizeFirst}[1]{
  \expandafter\@CapitalizeFirst\expandafter{#1}
}
\def\@CapitalizeFirst#1{\MakeUppercase{\@car#1\@nil}\@cdr#1\@nil}
\makeatother

\makeatletter
\DeclareRobustCommand\onedot{\futurelet\@let@token\@onedot}
\def\@onedot{\ifx\@let@token.\else.\null\fi\xspace}
\def\eg{\emph{e.g}\onedot} 
\def\ie{\emph{i.e}\onedot} 
 
 \def\vs{\emph{vs}\onedot}
 
\def\etal{\emph{et al}\onedot}

\makeatother

\renewcommand{\paragraph}[1]{\vspace{1pt}\noindent\textbf{#1}}

\definecolor{lightgray}{gray}{0.93}

%% file: tex/00_abstract.tex
\begin{abstract} 
Efficiently merging several models fine-tuned for different tasks, but stemming from the same pretrained base model, is of great practical interest. Despite extensive prior work, most evaluations of model merging in computer vision are restricted to image classification using CLIP, where different classification datasets define different tasks. In this work, our goal is to make model merging more practical and show its relevance on challenging scenarios beyond this specific setting.
In most vision scenarios,
different tasks rely on \textit{trainable} 
and usually \textit{heterogeneous} 
decoders.
Differently from previous studies with frozen decoders, 
 where merged models can be evaluated right away, the non-trivial cost of decoder training renders hyperparameter selection based on downstream performance impractical. To address this, we introduce the \emph{\taskalign}, and show how it can be used to speed up hyperparameter selection by orders of magnitude while retaining performance. Equipped with the \taskalign, we extend the applicability of model merging to multi-task vision models beyond CLIP-based classification.

 Project page: \url{https://europe.naverlabs.com/task-alignment}
    \keywords{Model Merging \and Task Alignment Proxy \and Merging LiDAR Models \and Heterogeneous Vision Tasks}
\end{abstract}

%% file: tex/01_intro.tex
\section{Introduction}
\label{sec:introduction}

\begin{figure}[t]
    \includegraphics[width=\linewidth]{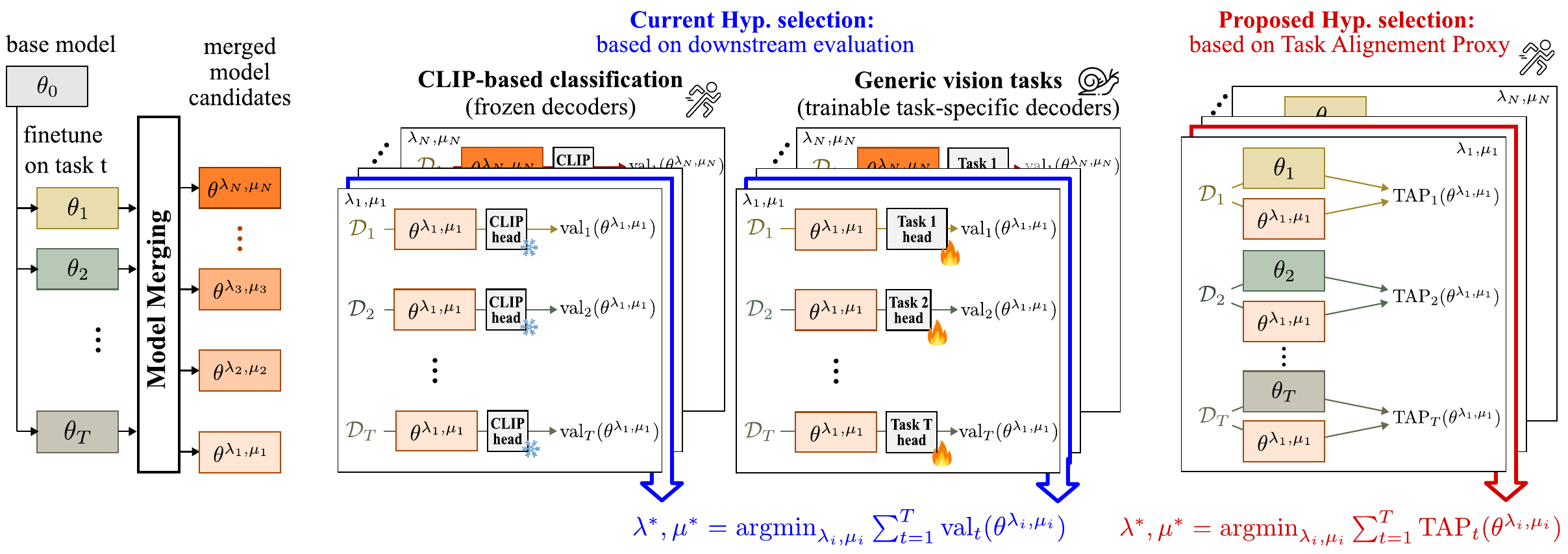}
    \caption{(Left) Model merging methods produce a large number of candidates which depend on hyperparameters $(\lambda,\mu)$. (Middle) SOTA methods select hyperparameters based on validation performance. 
    Evaluating and ranking all candidates might be feasible when the CLIP's text encoder is used as the frozen head (standard benchmarks) but becomes infeasible for more complex sets of tasks that require fine-tuning multiple task-specific decoders of different sizes, \textit{for each hyperparameter candidate}. (Right) Our proposed \TaskAlign (TAP) can be used to select hyperparameters at a fraction of the cost while maintaining performance (see also \cref{fig:cost_vs_perf}).}
    \label{fig:benchmark}
\end{figure}

Vision foundation models have emerged as generic encoders that support a wide range of 
computer vision tasks~\cite{oquab2024dinov2, alayrac2022flamingo, sariyildiz2024unic, sariyildiz2025dune, ranzinger2024radio, radford2021clip}, typically paired with task-specific decoders. Fine-tuning such models for individual tasks improves performance, but deploying multiple fine-tuned encoders is costly: each additional model increases memory and compute requirements. This is particularly problematic in resource-constrained settings such as robotics, where localization, human detection, and semantic or geometric scene understanding must run simultaneously on limited hardware.
This motivates a practical question: 
Can we efficiently merge multiple task-specialized encoders, fine-tuned from the same base, into a single encoder that retains performance while reducing deployment cost?

Recent works on model merging have brought us closer to answering that question.
Several approaches~\cite{ilharco2023editing, lee2025star, ortizjimenez2023tangent, wang2025lines,marczak2025notaskleftbehind} demonstrate that models fine-tuned from a shared initialization can be combined without additional training via simple arithmetic in the weight space. However, in computer vision, empirical evaluations have largely focused on CLIP-based image classification, where different label sets define different tasks. Because CLIP’s text encoder remains frozen, switching tasks only requires modifying class embeddings. This makes merged 
\input{floats/cost_vs_perf}
models easy and fast to evaluate, 
as no retraining is required.
Alternatively, some works studied 
merging on NLP tasks, where models have converged to being ``decoder-only'',
also allowing direct evaluation.

While encouraging, 
these settings obscure a fundamental challenge: in most vision applications, each task
relies on a different trainable decoder. 
In such scenario, merging can only be performed on part of the model, typically the shared encoder, while task-specific decoders must be fine-tuned after merging. 
This brings additional 
challenges and is the main focus of this paper, illustrated in~\cref{fig:benchmark}.

Relying on trainable decoders fundamentally changes the evaluation protocol. Unlike CLIP-based classification, where merged models can be assessed immediately, evaluating a merged model in this setting requires decoder training, a process that can be costly. As a result, exhaustive hyperparameter search over merging strategies becomes impractical, severely limiting applicability in most vision settings.

To address this bottleneck, we introduce the \TaskAlign (\talign), 
which efficiently evaluates merging candidates. \talign measures the alignment of feature spaces between models, enabling us to estimate merge compatibility without repeatedly fine-tuning decoders. By decoupling hyperparameter selection from expensive decoder training loops, \talign reduces the cost of model selection by orders of magnitude 
(models outlined in \textcolor{blue}{blue} vs in \textcolor{red}{red} in~\cref{fig:cost_vs_perf}).

We demonstrate that \talign enables model merging well beyond classification tasks. First, we show that in standard classification benchmarks, \talign accurately selects optimal merging configurations using only a few dozen unlabeled images (\cref{exp:clip}).
Second, we present, to the best of our knowledge, the first successful application of model merging to LiDAR. 
We focus on LiDAR point cloud semantic segmentation, with a single encoder being applied 
to diverse scenarios,
spanning different LiDAR sensors, environments, and annotated with different label sets. 
This demonstrates that merging generalizes beyond RGB (\cref{exp:lidar_merging}). 
Third, we apply \talign to the DUNE benchmark~\cite{sariyildiz2025dune} that considers a challenging multi-task scenario with
heterogeneous (\ie very diverse) tasks, namely 
semantic segmentation, depth estimation, 3D~pose relocalization, and human mesh recovery. 
In \cref{exp:dune_merging} we show \talign-guided merging can be applied even in this scenario to further boost the performance of a model obtained with large-scale multi-teacher distillation.
Notably,
since computing TAP does not require task-specific decoders, the same hyperparameter selection criterion (TAP) can be applied universally to merge models across CLIP classification, LiDAR segmentation, and heterogeneous 2D/3D vision settings; regardless of the set of downstream tasks.

%% file: floats/cost_vs_perf.tex
\begin{wrapfigure}[21]{r}{0.5\textwidth}
    \centering
    \vspace{-15pt}
    \includegraphics[width=0.48\textwidth]{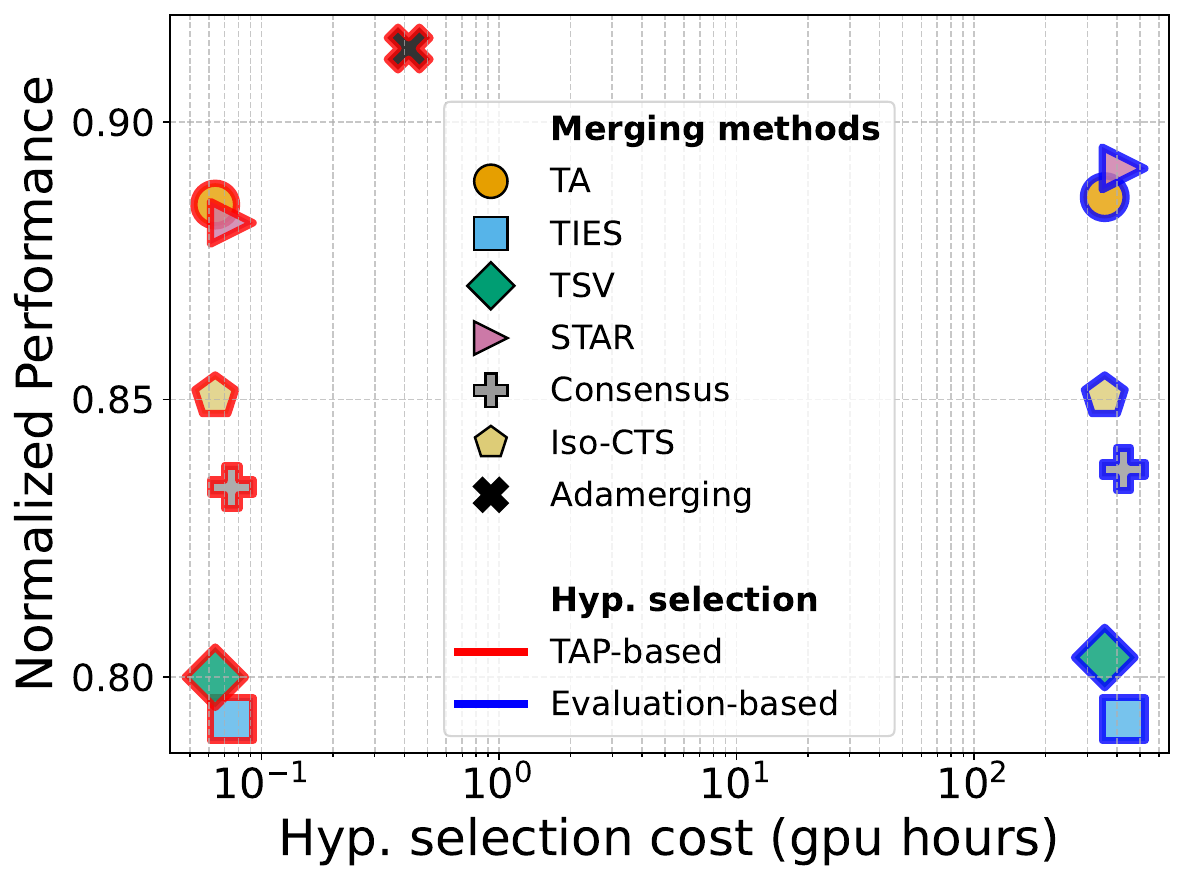}
    \caption{
      \textbf{Task Alignment Proxy (\talign) drastically reduces hyperparameter selection costs} when merging models for heterogeneous 2D and 3D vision tasks (see \cref{sec:protocol}). In this case, downstream evaluation is costly due to task-dependent decoder training, and hyperparameter search becomes impractical (\textcolor{blue}{blue border}). Hyperparameter selection with \talign reduces costs by 3 orders of magnitude (\textcolor{red}{red border}).
    }
    \label{fig:cost_vs_perf}
\end{wrapfigure}

%% file: tex/02_related_work.tex
\section{Background and related work}
\label{sec:related}

\paragraph{Preliminaries and notation.} Let $\theta_0 \in \mathbb{R}^d$ denote the parameters of a pretrained visual encoder, referred to as the \textit{base model}. We assume that this base model is fine-tuned end-to-end independently for each task $t$ in a set of $T$ downstream tasks. 
Let $\theta_t $ represent the encoder parameters after \finetuning for task $t$, defined as $\theta_t  = \argmin_\theta \mathcal{L}_t(\theta)$ where $\mathcal{L}_t(\theta)$ is the training loss for task $t$.

Following~\cite{ilharco2023editing}, we define the \textit{task vector} for task $t$ as $\tau_t = \theta_t  - \theta_0$. The goal of model merging is to find a single set of encoder parameters that jointly minimizes the losses across all tasks, \ie
$\theta^* = \operatorname*{argmin}_\theta \sum_{t=1}^T \mathcal{L}_t(\theta)$. \textit{Arithmetic-based merging}  methods aim to efficiently approximate $\theta^*$ by combining task vectors in the parameter space.
Most merging methods can be framed as a variation of the following weighted sum:
\begin{equation}
\label{eqn:th}
\theta_{\text{merged}}(\lambda, \mu) = \theta_0 + \sum_{t=1}^T \lambda_t  \odot \phi(\tau_t; \mu),
\end{equation}
where $\lambda_t  \in \mathbb{R}^d$ is a vector of task-specific weights,
$\phi: \mathbb{R}^d \rightarrow \mathbb{R}^d$ is a transformation parametrized by hyperparameters $\mu$ and $\odot$ is the Hadamard product. (\cref{app:related_work_extended} discusses how \cref{eqn:th} applies to different merging methods).
Let $\lambda$ denote the set of all $ \lambda_t $ parameters. When there is no ambiguity, we denote $\theta_{\text{merged}}(\lambda, \mu)$ as $\theta(\lambda, \mu)$ or just $\theta$. 

\paragraph{Model merging methods.}
The simplest approach is \textbf{model averaging}~\cite{matena2022merging, wortsman2022model, jin2023dataless, choi2024revisiting}, where $\lambda_t  = [1/T, ..., 1/T]$ and $\phi$ is the identity function. \textbf{Task Arithmetic}~\cite{ilharco2023editing} extends weight averaging by linearly interpolating between the average of fine-tuned models and the base model.
In \textbf{TIES}~\cite{prateek2023ties}, $\phi$ applies weight pruning to task vectors, retaining 
the values 
with highest magnitudes and consistent signs, in order to reduce task interference. \textbf{Breadcrumbs}~\cite{davari2023model} also applies pruning, but removes both high- and low-magnitude weights at the layer level to eliminate outliers. 
\textbf{Consensus}~\cite{wang2024localizing} prunes weights per task, then constructs a consensus mask by retaining weights that appear in more than one task-specific mask. 
\textbf{Lines}~\cite{wang2025lines} uses per-layer weights 
where coefficients increase linearly across layers, based on the intuition that early layers capture general features (thus closer to the base model) whereas later layers are more task-specific.

More recently, several methods apply SVD to per-layer weights, when weights are matrix-shaped, defaulting to Task Arithmetic otherwise. \textbf{STAR}~\cite{lee2025star} performs SVD truncation on each weight matrix of a task vector, retaining the top singular values to preserve a specified fraction of the energy, 
and rescales the remaining singular values to match the initial energy. \textbf{TSV}~\cite{gargiulo2025task} applies SVD twice: first independently to each task vector (as in STAR), and then again after concatenating the decomposed matrices across tasks to enforce orthogonality. \textbf{Iso}~\cite{marczak2025notaskleftbehind} applies a similar SVD-based strategy but balances per-task SVD with the SVD of the sum of task vectors with a tunable hyperparameter. Moreover, it imposes 
``isotropy''
by fixing all singular values to the average.

Finding optimal merging hyperparameters (lambdas, pruning and SVD truncation thresholds), although efficient for decoder-only NLP models or CLIP-based classification with frozen text embeddings, becomes prohibitively expensive when performing merging tasks that require \finetuning the decoders for evaluation, as is common in computer vision (\cref{fig:benchmark}).

Inspired by test-time adaptation, \textbf{AdaMerging}~\cite{AdaMerging_ICLR_2024} directly optimizes merging coefficients via entropy minimization. However, it is only applicable to CLIP-like settings with frozen classification heads. In \cref{sec:tas}, we extend it with our proposed \talign to \textit{any} set of tasks, but
the optimization requires storing all fine-tuned models in memory, which is very demanding as the number of tasks increases. 
\textbf{AdaMMS}~\cite{du2025adamms} was designed to merge LLMs and selects merging weights based on generation consistency between models obtained with adjacent hyperparameter values. Such requirements restrict it to methods with a single scalar hyperparameter, rendering it unsuitable for our setting.

In this work, we introduce the \taskalign and show it can \textit{accelerate hyperparameter search by several orders of magnitude} for any set of tasks.

\paragraph{Alternative model merging applications.}  Recently, some studies have explored merging in the context of continual learning.
Dziadzio \etal \cite{Dziadzio_2025_CVPR} use CLIP-based tasks in a continual learning setting where training data and inference objectives evolve over time.
Sokar \etal \cite{sokar2025continual} investigate model merging to mitigate catastrophic forgetting when \finetuning
Vision Language models on a continual VQA benchmark.
These studies find that simple methods (such as task arithmetic) perform surprisingly well, often matching or outperforming more complex approaches. Although our setup differs significantly, this is consistent with our observations in \cref{exp:lidar_merging} and \cref{exp:dune_merging}.

%% file: tex/03_TAS.tex
\section{Task alignment} 

As discussed before, most model merging evaluations in vision have focused on CLIP-based zero-shot image classification.
In this section, we ask: \textit{Can we apply model merging to \textit{any} set of vision tasks?}
We first discuss the new challenges when merging models with trainable decoders (see  \cref{sec:partial_merging}). This motivates our generic and efficient performance proxy to guide hyperparameter selection in more complex scenarios (\cref{sec:tas}).

\subsection{Merging with trainable decoders} 
\label{sec:partial_merging}

The dominant evaluation setup for model merging in vision builds on CLIP's zero-shot classification capabilities (see~\cref{fig:benchmark}, left):  
the visual encoder is fine-tuned separately for each task, while the frozen text encoder serves as a shared 
classification head
(\ie a very light-weight decoder) by embedding class names or descriptions into a compatible feature space~\cite{radford2021clip,ilharco2021openclip}. As a result, the visual encoder effectively constitutes the \textit{entire model} used to solve all tasks. 
Hence, in this setting, evaluation of merged models is cheap and requires only forward inference,
making exhaustive search over merging hyperparameters 
feasible. 

However, many practical multi-task vision systems produce outputs in different spaces (\eg semantic labels, depth maps, 3D poses, meshes) and therefore rely on a task-specific decoder attached to an encoder. 
In this setting, only encoder parameters can be merged, while decoders remain task-specific and must be fine-tuned after merging.
Unlike CLIP-based classification, where merged encoders can be evaluated immediately by swapping label embeddings, here, each candidate encoder to merge $\theta(\lambda,\mu)$ must be paired with a newly fine-tuned decoder to assess performance. Fine-tuning decoders 
often requires hours, if not days, of compute per configuration, 
rendering exhaustive hyperparameter search intractable. This evaluation bottleneck severely limits the practical applicability of model merging across 
multi-task vision settings. 

One way to bring this setup closer to the CLIP-based formulation would be to keep decoders frozen during hyperparameter selection. However, this still requires the use of multiple, potentially expensive and complex decoders and oftentimes leads to suboptimal performance (see \cref{app:frozen_heads}).

Another possible solution is to use \textit{proxy} signals to efficiently estimate downstream performance.
Proxy-based approaches \adamerge~\cite{AdaMerging_ICLR_2024} and AdaMMS~\cite{du2025adamms} follow this direction. However, \adamerge relies on entropy minimization, which is only applicable to categorical tasks such as classification, while AdaMMS is specifically designed for generative models and assumes one can compare generated outputs. These limitations motivate us to introduce a generic proxy measure, applicable across \textit{any} set of tasks, enabling both efficient hyperparameter selection and learning-based merging.

\subsection{\TaskAlign} 
\label{sec:tas}

Motivated by the prohibitive cost of hyperparameter tuning when tasks require decoder training,
we propose a generic proxy 
to approximate
the performance drop incurred by swapping, for a given task $t$,
the fine-tuned parameters $\theta_t$ with the merged model $\theta_\text{merged}$. 
We refer to this as the \textbf{\TaskAlign} (\textbf{\talign}).
The intuition behind \talign 
is the following: 
if the features of the fine-tuned encoder $\theta_t$ and the merged encoder $\theta_\text{merged}$ are similar for images of task $t$, then downstream performance should also be similar for this task.
Computing \talign only requires access to {\em unlabeled} data from each downstream task and is both task-agnostic and computationally efficient, as it only involves inference.

Formally, let $D_1, \dots, D_T$ denote unlabeled training image sets for each task.
Let $f(x; \theta): \mathcal{X} \rightarrow \mathcal{F}$ denote the image encoder parameterized by $\theta$, where $\mathcal{X}$ is the input space and $\mathcal{F}$ the output feature space, and let $x \in \mathcal{X}$. Given a dissimilarity function $d: \mathcal{F} \times \mathcal{F} \rightarrow \mathbb{R}^+$, we define the \emph{\taskalign} for task $t$ as the average dissimilarity between features produced by the merged model and the task-specific fine-tuned model over $N$ samples $x_i \sim D_t$: 
\begin{equation}
    \text{\talign}_{t} (\theta) = \dfrac{1}{N} \sum_{i=1}^N d \big( f(x_i; \theta), f(x_i; \theta_t) \big), \ \ x_i \sim D_t
    \label{eq:talign}
\end{equation}
To obtain a single score across all tasks, 
we average over the tasks and compute
$\text{\talign}(\theta) = \sum_{t=1}^T \text{\talign}_t(\theta)/T$. 

\noindent \textbf{Default settings.} We use the $\ell_2$ distance for $d$ and 128 samples for $N$. Yet, in \cref{sec:tas_ablations} we show that \talign is robust to the choice of dissimilarity function $d$ and the number of samples used $N$.

\begin{figure}[t]
  \centering
  \includegraphics[width=0.99\textwidth]{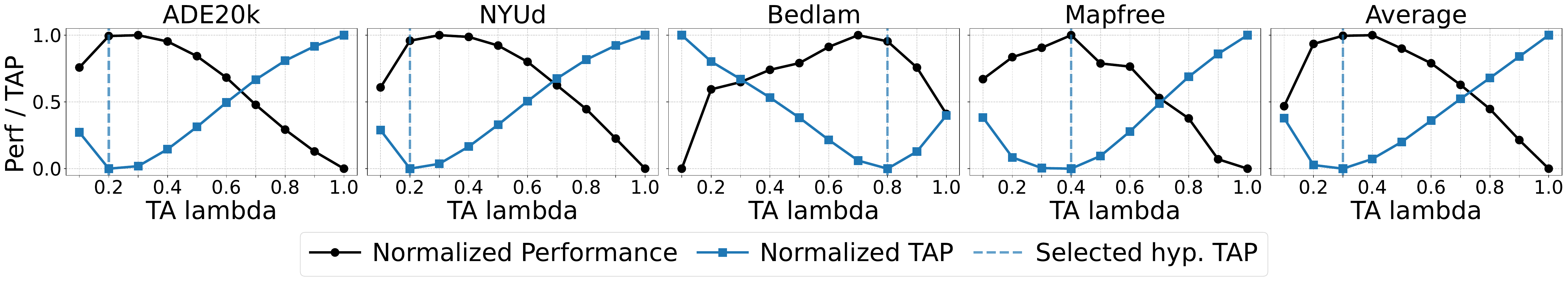}
    \caption{
    \textbf{Normalized \talign \vs performance} for different values of the merging coefficient $\lambda$ when using Task Arithmetic (TA) on the heterogeneous task setting, see \cref{sec:protocol}.
    The selected value is indicated by a dashed line. 
    }
    \label{fig:task_alignment}
\end{figure}

\input{floats/tas_vs_performance}

\paragraph{Combining \talign with existing merging methods.}
In the following sections, we combine \talign with several merging methods to select hyperparameters. We denote such combinations with an additional ``{\scriptsize w/\talign}'' (\eg TSV$_\text{w/\talign}$).
This is done by selecting  
the $\lambda$ (and whenever relevant $\mu$) parameters that minimize 
\talign, \ie $\lambda^*,\mu^*$ = $\argmin_{\lambda, \mu} \text{\talign}(\theta_\text{merged}(\lambda,  \mu))$, where $\theta_\text{merged}(\lambda,  \mu)$ is given by \cref{eqn:th}. 
We specify the standard hyperparameter selection based on downstream performance evaluation with ``{\scriptsize w/\HS}''; see \cref{fig:benchmark} for an illustration.

While most methods rely on hyperparameter search for method selection, \adamerge finds the optimal $\lambda$ via entropy minimization of the merged model on the different task datasets. As discussed in \cref{sec:related}, this cannot be directly applied to non-categorical tasks, \ie beyond classification. To extend \adamerge to any set of downstream tasks, we replace the entropy loss with \talign, which can also be treated
as a loss, and we name this version AdaMerging$_\text{w/\talign}$.

%% file: floats/tas_vs_performance.tex

\begin{figure}[t]
  \centering
  \setlength{\tabcolsep}{3pt}  
  \renewcommand{\arraystretch}{0} 

  \begin{tabular}{ccc}
    \multicolumn{1}{c}{~~~~\scriptsize CLIP classification} &
    \multicolumn{1}{c}{~~~~\scriptsize LiDAR segmentation} &
    \multicolumn{1}{c}{~~~~\scriptsize Heterogeneous tasks} \\[4pt]
    \includegraphics[width=0.31\linewidth]{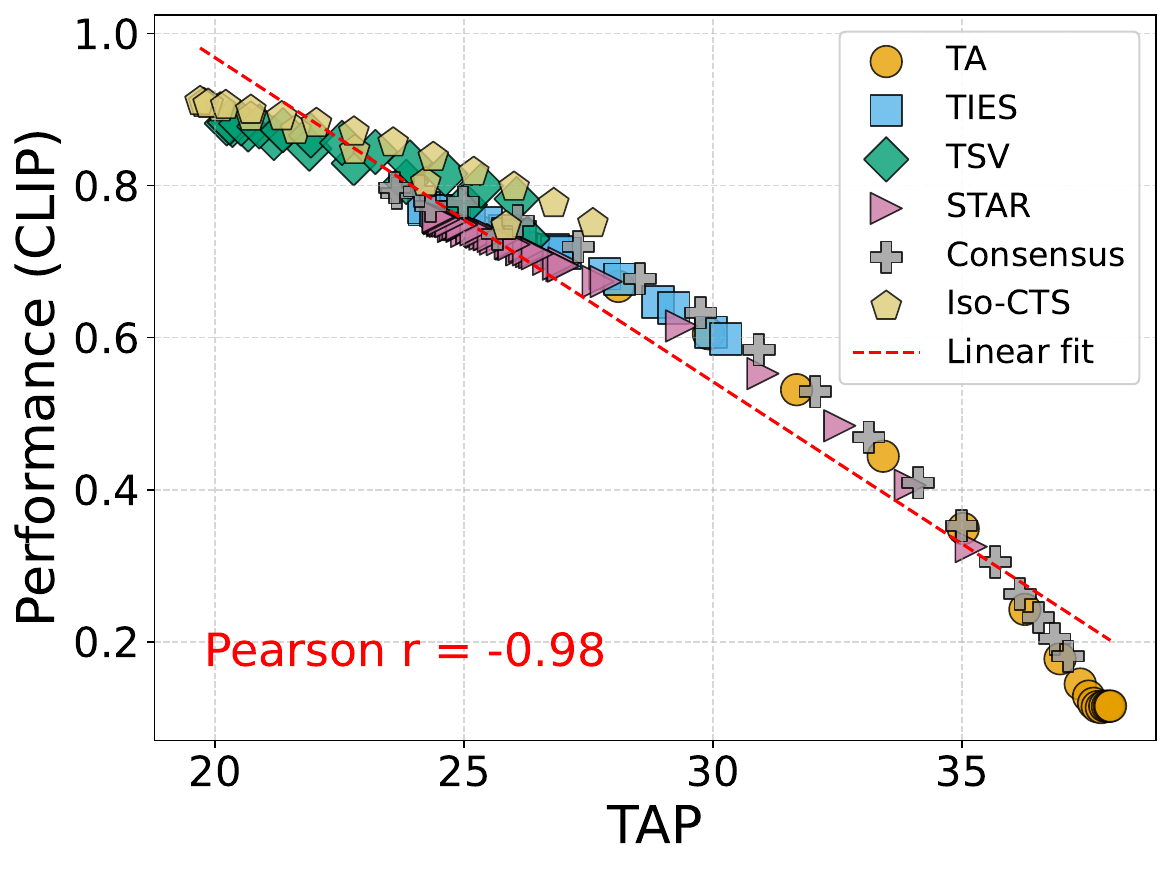} &
    \includegraphics[width=0.31\linewidth]{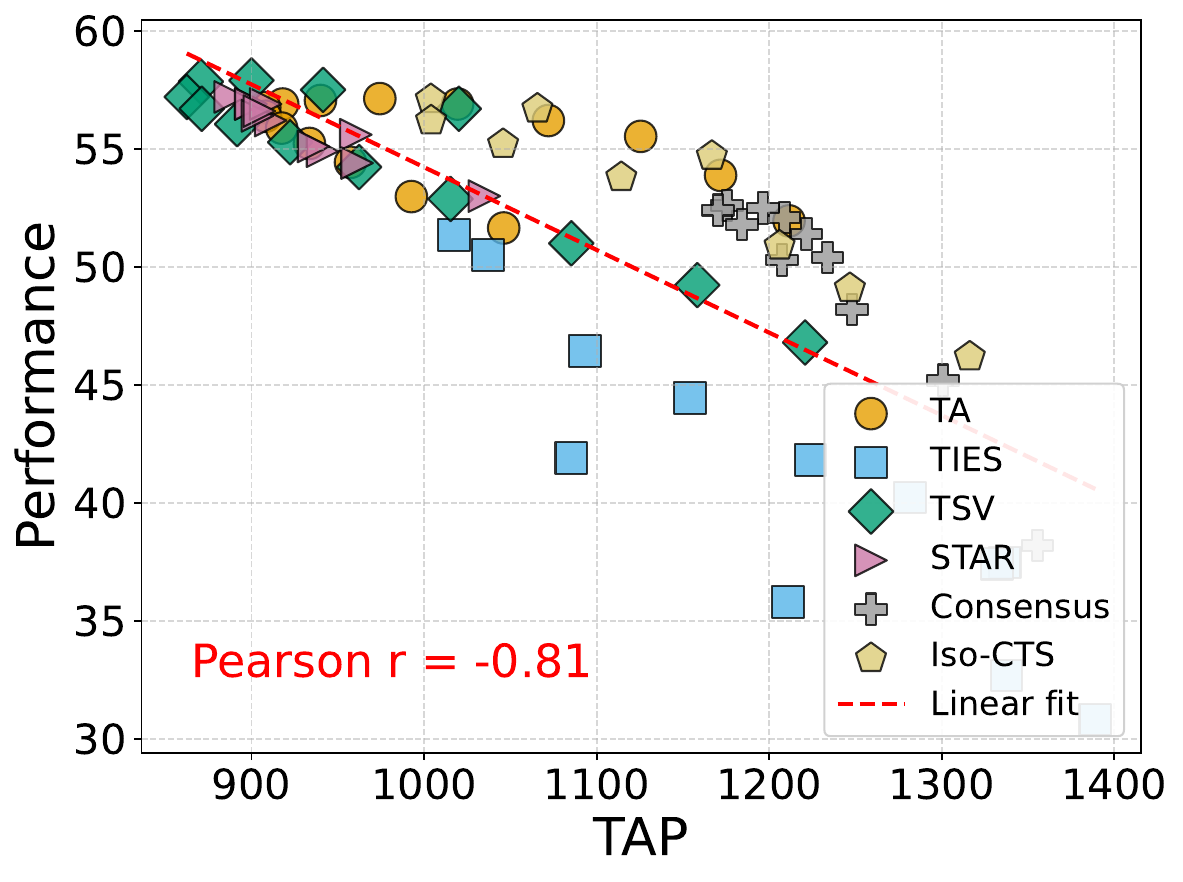} &
    \includegraphics[width=0.31\linewidth]{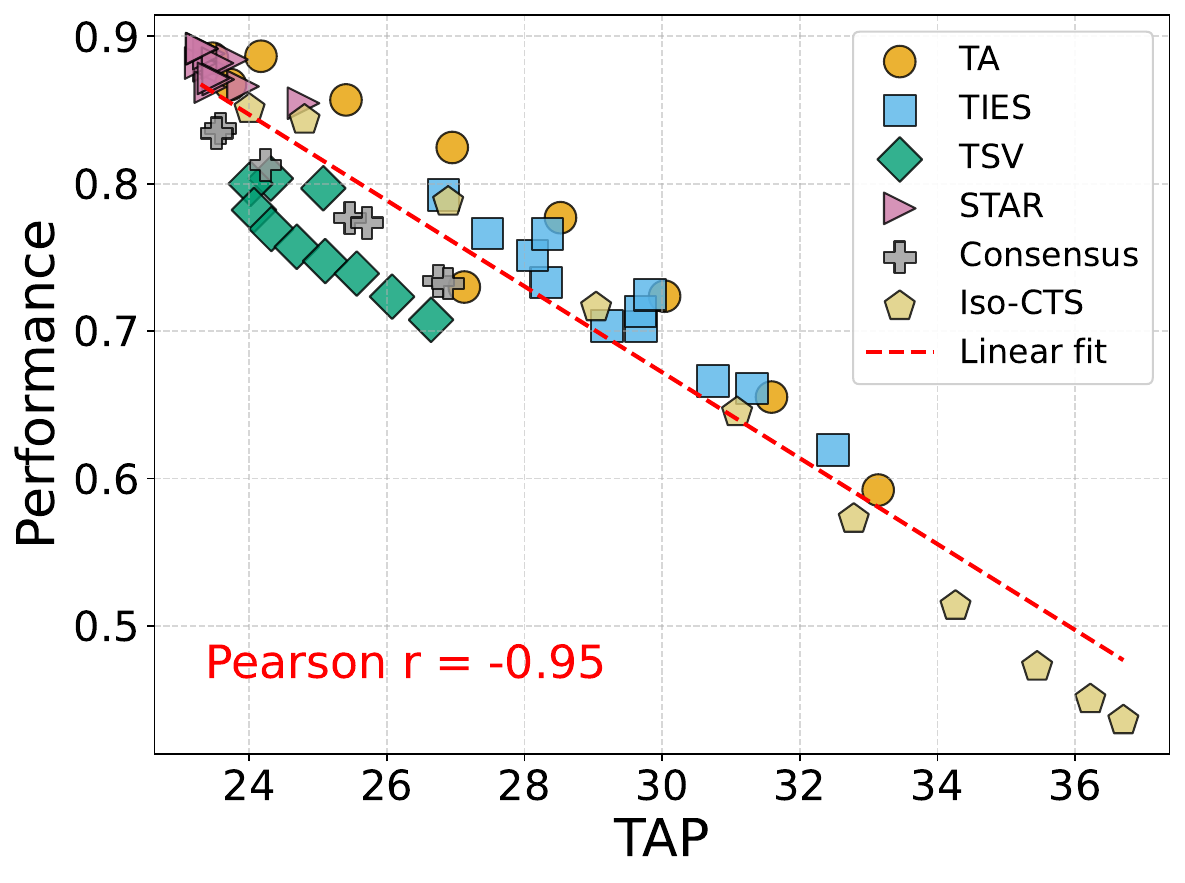} \\ [4pt]
    \includegraphics[width=0.31\linewidth]{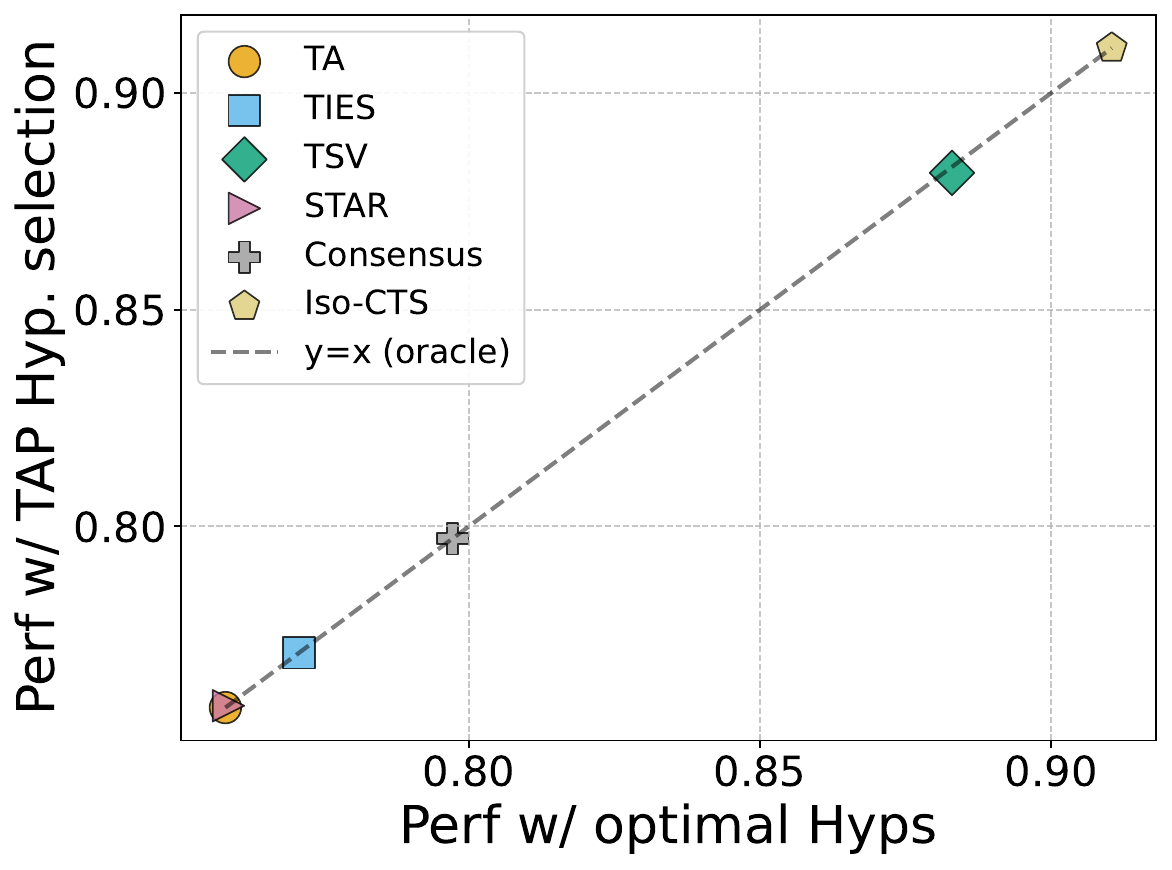} &
    \includegraphics[width=0.31\linewidth]{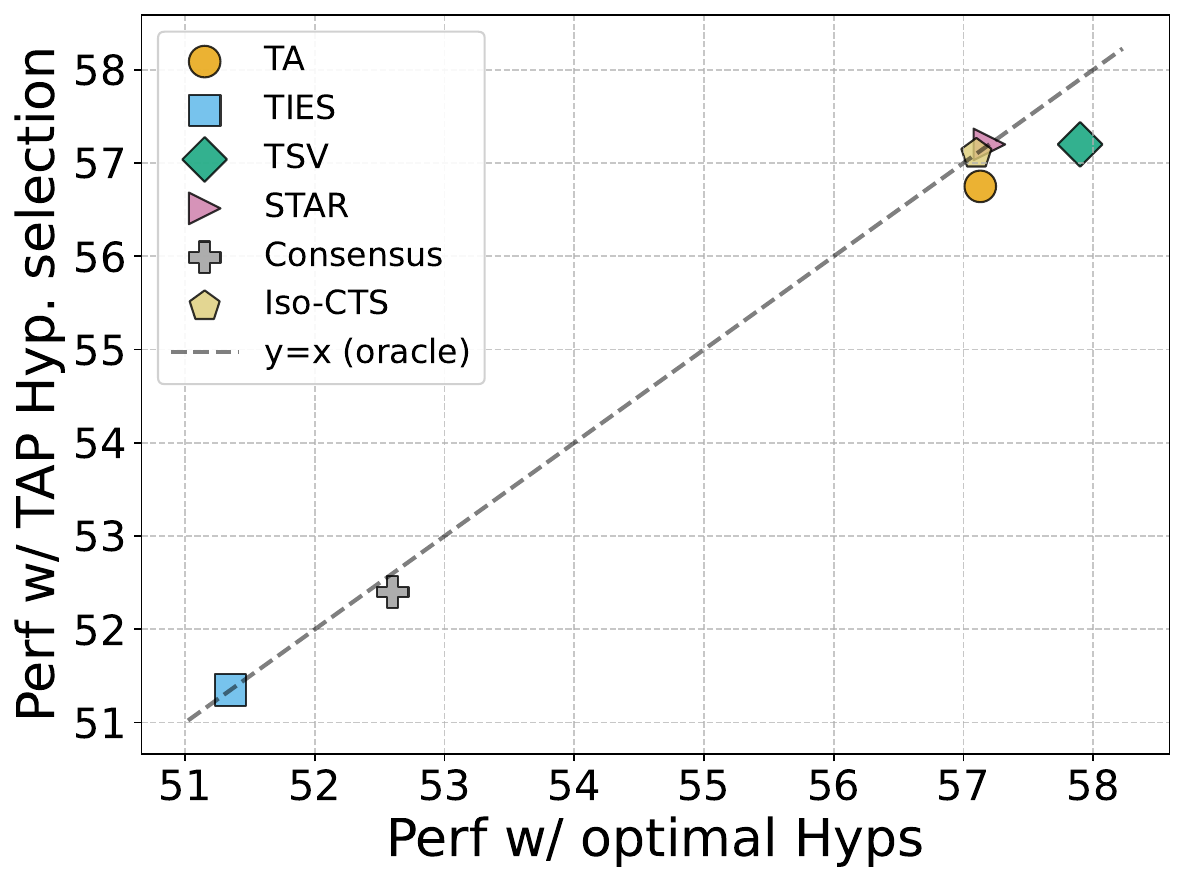} &
    \includegraphics[width=0.31\linewidth]{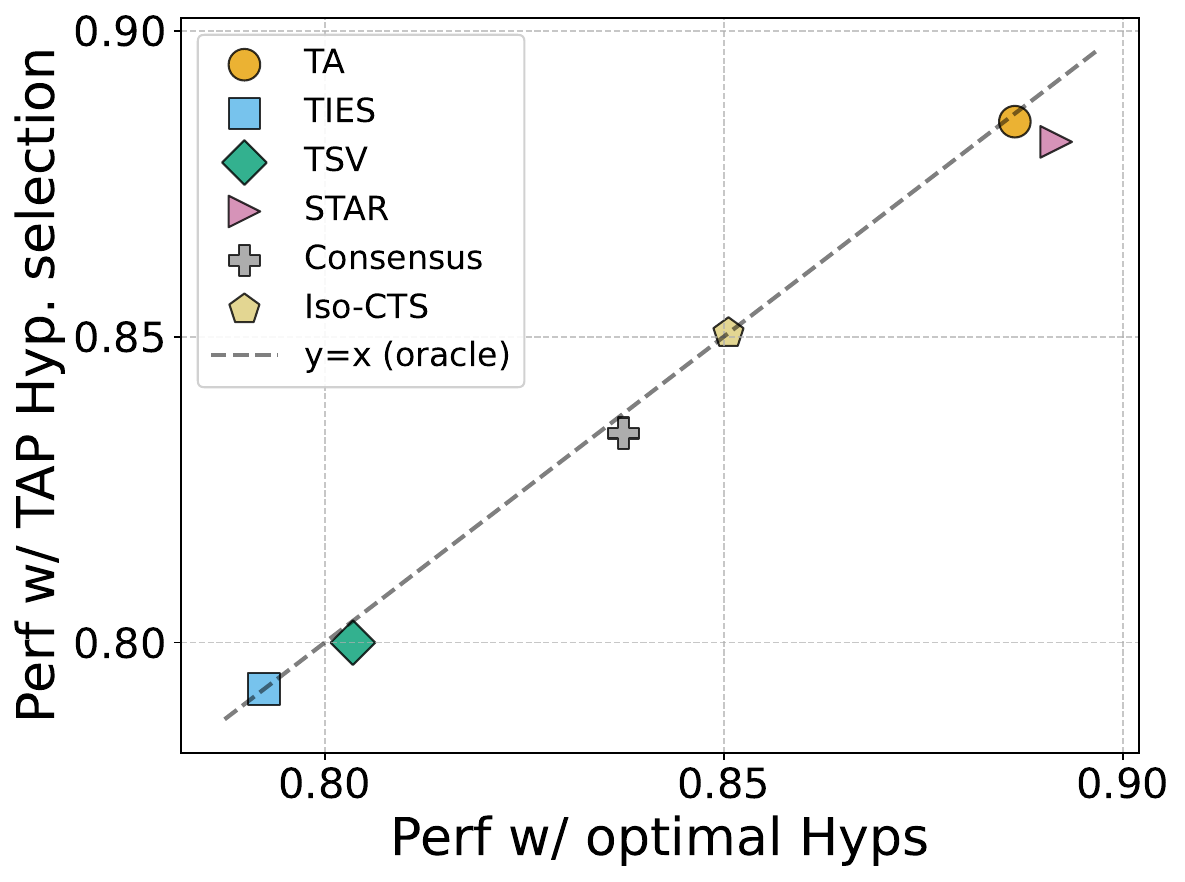} \\
  \end{tabular}
  \caption{
  \textbf{Top row:} Correlation between \TaskAlign (TAP) and Performance on the validation sets when merging models for CLIP classification with a ViT-L: 20 task setting (left), 3D segmentation with different LiDAR sensors (middle) and heterogeneous vision tasks (right).
  \textbf{Bottom row:} Performance with \talign-selected hyperparameters \vs optimal hyperparameters (with costly downstream evaluation) for the same settings. We observe a strong correlation between \talign and performance which leads to almost optimal hyperparameters \textit{without any downstream task evaluation}.
  %
  }
  \label{fig:talign_scatters}
\end{figure}

%% file: tex/05_results.tex
\section{Experiments}
\label{sec:experiments}

\subsection{Experimental protocol}
\label{sec:protocol}
We perform model merging in three main settings: CLIP-based classification on multiple datasets, LiDAR-based segmentation with different sensors and data domains, and a combination of heterogeneous vision tasks introduced in~\cite{sariyildiz2025dune}.

\paragraph{CLIP-based classification.} Initially introduced in \cite{ilharco2023editing} and later re-used and extended in many works, \cite{wang2024localizing,gargiulo2025task,AdaMerging_ICLR_2024,prateek2023ties,wang2025lines,marczak2025notaskleftbehind} to name a few, this has been the de-facto evaluation protocol for model merging in vision. We follow the evaluation code and checkpoints from \cite{gargiulo2025task} and evaluate merging across three different architectures (ViT-B with patch size 32, ViT-B with patch size 16 and ViT-L with patch size 14) and incremental sets of tasks with 8, 14 and 20 datasets respectively. For more details we refer the reader to \cite{gargiulo2025task}.

\paragraph{LiDAR segmentation.}
We consider four different 3D LiDAR semantic segmentation datasets: 
nuScenes~\cite{caesar2020nuscenes,fong2021lidarseg_nuscenes}, SemanticKITTI~\cite{geiger2012cvpr, behley2019iccv}, Panda64~\cite{xiao2021pandaset} and PandaGT~\cite{xiao2021pandaset}; captured with different LiDAR sensors (\eg varying number of beams, resolution or field of view) and different label taxonomy.
We use the WaffleIron-768 (WI) architecture~\cite{puy23waffleiron} as an encoder,
pretrained with ScaLR~\cite{puy2024three}, a multi-modal image-to-LiDAR distillation, and use linear decoders. 

Regarding per-task fine-tuning, we use the models from~\cite{puy2024three} as they are the highest performing models publicly available at the time of writing.
Once fine-tuned encoders are merged, 
an independent decoder is trained for each task on top of the frozen encoder. After decoder training, we evaluate the performance on each task following the protocol in~\cite{puy2024three}.

\paragraph{Heterogeneous vision tasks.} We adopt the evaluation protocol introduced in DUNE~\cite{sariyildiz2025dune} to assess model merging across a diverse set of 2D and 3D vision tasks; we refer to it as the \textbf{DUNE benchmark}. Specifically, we evaluate on semantic segmentation using \ade~\cite{zhou2019semantic}, depth estimation on NYUd~\cite{nyu}, 3D human mesh recovery on Bedlam~\cite{bedlam}, and pose relocalization on Niantic's \mapfree~dataset~\cite{mapfree}. As a base model we use DUNE (ViT-Base/14), given its demonstrated performance on such a broad range of tasks,
precisely the setting we want to study. 
We also start from DINOv2 as a base model in that particular setting and report results in \cref{sec:dinov2_merging}.

We fine-tune the base model end-to-end for each task, resulting in four task-specific encoders. After merging, we freeze the encoder and fine-tune only the corresponding task-specific decoders. For the decoders, we use a linear head for semantic segmentation (\ade) and a DPT model for depth estimation (\nyud), following~\cite{oquab2024dinov2}. For human mesh recovery on \bedlam, we adopt the two-headed architecture from MultiHMR~\cite{multihmr}. For visual relocalization on \mapfree, we use a ViT-Large decoder as proposed in MASt3R~\cite{mast3r}. 

We follow the evaluation of DUNE, so results reported in \cref{tab:dune} and \cref{tab:dinov2} are directly comparable to those in~\cite{sariyildiz2025dune}. However, due to the high cost of model evaluation, we introduce an \textit{ablation} protocol which is used when evaluating different hyperparameters per model in \cref{fig:talign_scatters}. In this protocol, we reduce the number of \finetuning iterations for task-specific decoders, 
and, in the case of \mapfree, we also limit the number of training samples.
Moreover, we use a linear head (instead of DPT) for depth estimation. To avoid bias towards the original validation sets, evaluations are conducted on custom held-out sets. 
Further details on our experimental settings can be found in \cref{app:experimental_details}.

\subsection{Is \talign a good proxy for performance?}

In this section we show that \talign 
~strongly correlates
with downstream performance and that \talign selection is comparable to that of using an oracle (\ie selecting parameters based on full downstream evaluation).


In \cref{fig:task_alignment} we merge the different models in the heterogeneous task setting with Task Arithmetic (TA). We consider different hyperparameter values and show \talign (in blue) \vs the actual performance after the costly decoder \finetuning per-task (in black), for each task (first four plots) and averaged across tasks (right-most plot).
In both cases, \ie per-task \talign and average \talign,
we observe a strong correlation between 
 performance and \talign.

\cref{fig:talign_scatters} (top) shows the average normalized performance \vs the average \talign for different merging methods and hyperparameter settings.
\cref{fig:talign_scatters} (bottom) shows the performance of the models selected with \talign \vs the actual best model for each method found via exhaustive evaluation of the
hyperparameter configurations. 
We report these results for the three different settings described in \cref{sec:protocol}. As already noted in \cref{fig:task_alignment}, we 
observe a strong negative correlation between \talign and performance. Moreover, on the bottom plots we observe that all methods are close to the $y=x$ line;
that corresponds
to always selecting  
hyperparameters leading to the best performance.
Hence, \talign is indeed a good proxy that can be used for hyperparameter selection across a variety of settings in computer vision. In the following sections we present more detailed results for each merging setting.

\input{tables/clip_summary_2}

\subsection{CLIP-based image classification}
\label{exp:clip}

In this section we compare the performance of evaluation \vs \talign-based hyperparameter selection in the standard CLIP-based image classification benchmark. In \cref{tab:clip_summary} we report the test performance for different merging methods when selecting hyperparameters following one of the two strategies, 
across different architectures and task sets. We observe that \talign-based selection leads 
to comparable performance to exhaustive hyperparameter selection.
We even see \talign selection achieving higher test performance (by a small margin) in some cases.
We attribute these small variations to validation-test set noise. 

For \adamerge, there  seems to be a trend where optimizing \talign leads to better performance than with the original entropy-based formulation as we increase the number of tasks and architecture size.

It is also worth noting, that, although exhaustive hyperparameter evaluation in this setting is feasible thanks to the use of frozen CLIP decoders, computing \talign does not require labels. Moreover, in \cref{fig:tas_bs_ablation} we show that \talign is very stable even when using very few images.
This shows that \talign-based selection would still be beneficial in cases where labels are not available, or only a few samples per-task are kept, \eg due to privacy or memory constraints.

\subsection{Merging LiDAR models}
\label{exp:lidar_merging}

\input{tables/lidar_merging}

To illustrate the broad applicability of model merging in computer vision, we merge different models fine-tuned to perform LiDAR-based semantic segmentation.
When using LiDAR inputs, besides potential domain gaps due to geolocation or weather conditions (as one would encounter in RGB-based vision),
changes in the LiDAR sensors used to acquire the data often lead to stark drops in performance~\cite{yi2021complete, michele2024ttyd}. Hence, it is quite common to start from a strong pretrained model and fine-tune it for the relevant sensor~\cite{puy2024three}. However, if inputs from different sensors
need to be processed simultaneously, \eg on a robot with multiple LiDAR sensors, one would need to deploy several models at once. 

In \cref{tab:lidar}, we show that model merging (with \talign-based hyperparameter selection) can be used to effectively merge different encoders fine-tuned on different datasets/sensors, significantly improving the performance over the pretrained model. To the best of our knowledge, this is the first work to show that model merging can be successfully applied to LiDAR models. To avoid redundancy, we only report numbers with TAP-based hyperparameter selection, but a table with evaluation-based selection can be found in \cref{app:further_exps_lidar}.

Similar to the findings in \cref{fig:cost_vs_perf} for heterogeneous tasks, \talign accelerates hyperparameter selection in this setting significantly. While decoder training requires more than 30 gpu hours for each evaluated hyperparameter, computing \talign requires approximately only 7 minutes, \textit{accelerating hyperparameter selection by two orders of magnitude}. For further details see \cref{app:further_exps_lidar}.

\subsection{Merging models for heterogeneous vision tasks}
\label{exp:dune_merging}

\input{tables/dune_merging}

In this section we study the effectiveness of model merging when downstream tasks (and decoders) are significantly different. We would like to note that, given the task diversity, it is not obvious that such models could be merged effectively, even with expensive merging techniques such as distillation, as discussed in \cite{sariyildiz2025dune}. In \cref{tab:dune} we apply different model merging methods with \talign-based hyperparameter selection and show they can indeed improve the performance of a state-of-the-art multi-task vision model, \ie DUNE\cite{sariyildiz2025dune}, via further fine-tuning (on top of DUNE) and merging. To the best of our knowledge, no prior work has evaluated model merging on a set of heterogeneous vision tasks spanning both 2D and 3D. 
This is the setting where TAP speeds up hyperparameter selection the most, and this is mainly due to the expensive decoder training protocol of the MASt3R decoder \cite{mast3r} (used for MapFree).
\textit{As illustrated in \cref{fig:cost_vs_perf}, we can reduce hyperparameter search by three orders of magnitude}. Due to its prohibitive cost, we do not perform evaluation-based hyperparameter selection in this setting using the original 
decoder training protocol from \cite{sariyildiz2025dune}.

\input{floats/BS_ablation}

%% file: tables/clip_summary_2.tex
\begin{table}[t]
\centering
\caption{\textbf{Comparing hyp. selection strategies on CLIP benchmark.} Test performance of models selected via standard hyp. selection (``{\scriptsize w/\HS}'') or using \talign on a subset of 128 unlabeled images per task (``{\scriptsize w/\talign}''). Both 
lead to comparable performance across merging methods and settings. We report ViT-B-16 in \cref{app:further_exps_clip}. \textit{Fine-tuned} refers to an upper-bound for which separate expert models are trained per task.}
\label{tab:clip_summary}
\setlength{\tabcolsep}{3.5pt}
\begin{tabular}{l p{3pt} r r r p{3pt} r r r r}
\toprule
& & \multicolumn{3}{c}{\texttt{ViT-B-32}} 
& & \multicolumn{3}{c}{\texttt{ViT-L-14}} 
& \multirow{2}{*}{Average} \\
\cmidrule(lr){3-5}
\cmidrule(lr){7-9}
& & T=8 & T=14 & T=20 
& & T=8 & T=14 & T=20 
& \\
\midrule
Zero-shot & & 48.3 & 57.2 & 56.1 & & 64.7 & 68.2 & 65.2 & 60.0 \\
Fine-tuned & & 92.8 & 90.9 & 91.3 & & 95.8 & 94.3 & 94.7 & 93.3 \\
\midrule
\rowcolor{gray!10}
TA$_{\text{w/\HS}}$      & & 71.1 & 65.2 & 62.6 & & 84.9 & 79.4 & 76.0 & 72.7 \\
\rowcolor{gray!10}
TA$_{\text{w/\talign}}$  & & 71.1    & 65.0 & 62.1 & & 84.9 & 79.7 & 76.0 & 72.5 \\

\addlinespace[0.7ex]

STAR$_{\text{w/\HS}}$    & & 71.4 & 65.4 & 62.9 & & 84.9 & 79.7 & 76.1 & 72.9 \\
STAR$_{\text{w/\talign}}$& & 70.9 & 65.3 & 63.0 & & 84.9 & 79.7 & 76.1 & 72.8 \\

\addlinespace[0.7ex]
\rowcolor{gray!10}
TIES$_{\text{w/\HS}}$    & & 74.8 & 67.8 & 64.8 & & 86.9 & 80.2 & 77.1 & 74.9 \\
\rowcolor{gray!10}
TIES$_{\text{w/\talign}}$& & 74.5 & 67.8 & 64.8 & & 87.0 & 80.2 & 77.1 & 74.9 \\

\addlinespace[0.7ex]

Consensus$_{\text{w/\HS}}$   & & 74.9 & 70.3 & 66.9 & & 86.3 & 82.3 & 79.6 & 76.2 \\
Consensus$_{\text{w/\talign}}$& & 74.7 & 70.3 & 66.9 & & 86.3 & 82.3 & 79.6 & 76.2 \\

\addlinespace[0.7ex]

\rowcolor{gray!10}
TSV$_{\text{w/\HS}}$     & & 85.8 & 80.0 & 76.9 & & 93.0 & 89.2 & 87.5 & 85.2 \\
\rowcolor{gray!10}
TSV$_{\text{w/\talign}}$ & & 85.5 & 80.0 & 76.9 & & 93.0 & 89.1 & 87.6 & 85.1 \\

\addlinespace[0.7ex]

Iso-CTS$_{\text{w/\HS}}$ & & 86.5 & 81.6 & 78.2 & & 94.8 & 91.0 & 90.2 & 86.9 \\
Iso-CTS$_{\text{w/\talign}}$ & & { 85.8} & 81.6 & { 78.1} & & { 94.7} & 91.0 & 90.2 & { 86.8} \\

\addlinespace[0.7ex]

\rowcolor{gray!10}
AdaMerging$_{\text{w/Entropy}}$ & & 84.2 & 75.0 & 71.5 & & 91.4 & 87.0 & 84.3 & 81.1 \\
\rowcolor{gray!10}
AdaMerging$_{\text{w/\talign}}$  & & 80.7 & 75.4 & 72.4 & & 91.7 & 87.7 & 87.1 & 81.5 \\

\bottomrule

 











\end{tabular}
\end{table}

%% file: tables/lidar_merging.tex
\begin{table}[t]
\caption{
\textbf{Merging LiDAR models.} Results after merging different 3D segmentation models with TAP-based hyperparameter selection. We successfully apply several methods, significantly improving over the pretrained baseline. We report the mIoU and the normalized performance w.r.t. the fine-tuned models.
}
\label{tab:lidar}
\centering
\begin{tabular*}{\textwidth}{@{\extracolsep{\fill}}lrrrrr}
\toprule
\makecell{ \\ } & \makecell{nuScenes \\ mIoU (\(\uparrow\))} & \makecell{Sem.KITTI \\ mIoU (\(\uparrow\))} & \makecell{Panda64 \\ mIoU (\(\uparrow\))} & \makecell{PandaGT \\ mIoU (\(\uparrow\))} & \makecell{Normalized \\ Performance} \\
\midrule
Pretrained & 68.4 & 55.6 & 37.0 & 35.4 & 0.841 \\
\midrule
TA$_{\text{w/\talign}}$ & 73.7 & 63.5 & 47.0 & 42.9 & 0.972 \\
STAR$_{\text{w/\talign}}$ & 74.1 & {63.7} & \textbf{47.5} & 43.3 & {0.979} \\
TIES$_{\text{w/\talign}}$ &
69.0 & 56.0 & 41.9 & 38.5 & 0.879 \\
Consensus$_{\text{w/\talign}}$ & 70.1 & 59.5 & 41.3 & 38.8  & 0.898\\
TSV$_{\text{w/\talign}}$ & 74.7 & 63.5 & 47.2 & 43.4 & {0.979} \\
Iso-CTS$_{\text{w/\talign}}$ & \textbf{74.8} & 62.4 & 47.0 & 44.6 & 0.979 \\
AdaMerging$_{\text{w/\talign}}$ & 73.1 & \textbf{64.4} & 46.6 & \textbf{46.1} & \textbf{0.985} \\
\bottomrule
\end{tabular*}
\end{table}

%% file: tables/dune_merging.tex
\begin{table}[t]
\caption{
\textbf{Improving heterogeneous 
vision models via merging.} 
Several models merged with \talign-based hyperparameter selection outperform the pretrained model DUNE. 
$^\dagger$ denotes 
our reproduction of the public DUNE model results
under the exact same evaluation setting.
    }
\label{tab:dune}
\centering
\begin{tabular*}{\textwidth}{@{\extracolsep{\fill}}lrrrrr}
\toprule
\makecell{ \\ } & \makecell{\ade \\ mIoU (\(\uparrow\))} & \makecell{\nyud \\ rmse (\(\downarrow\))} & \makecell{\bedlam \\ pa-pve (\(\downarrow\))} & \makecell{\mapfree \\ AUC (\(\uparrow\))} & \makecell{Normalized \\ Performance} \\
\midrule
DUNE$^\dagger$ (Pretrained) & 45.6 & 0.330 & 62.5 & 94.7 & 0.929 \\
\midrule
TA$_{\text{w/\talign}}$ & 46.6 & 0.315 & 59.7 & 95.2 & 0.956 \\
STAR$_{\text{w/\talign}}$ & 46.6 & 0.312 & \textbf{59.3} & \textbf{95.7} & 0.962 \\
TIES$_{\text{w/\talign}}$ & 46.3 & 0.324 & 63.3 & 94.7 & 0.934 \\
Consensus$_{\text{w/\talign}}$ & 46.7 & 0.324 & 65.7 & 94.7 & 0.928 \\
TSV$_{\text{w/\talign}}$ & 45.6 & 0.321 & 62.9 & 95.5 & 0.936 \\
Iso-CTS$_{\text{w/\talign}}$ & 48.2 & 0.326 & 65.5 & 94.4 & 0.934 \\
AdaMerging$_{\text{w/\talign}}$ & \textbf{49.3} & \textbf{0.300} & 62.9 & 95.2 & \textbf{0.971} \\
\bottomrule
\end{tabular*}
\end{table}

%% file: floats/BS_ablation.tex
    
    

\begin{figure}[t]
    \centering
    \begin{minipage}{0.47\textwidth}
        \centering
        \includegraphics[width=\linewidth]{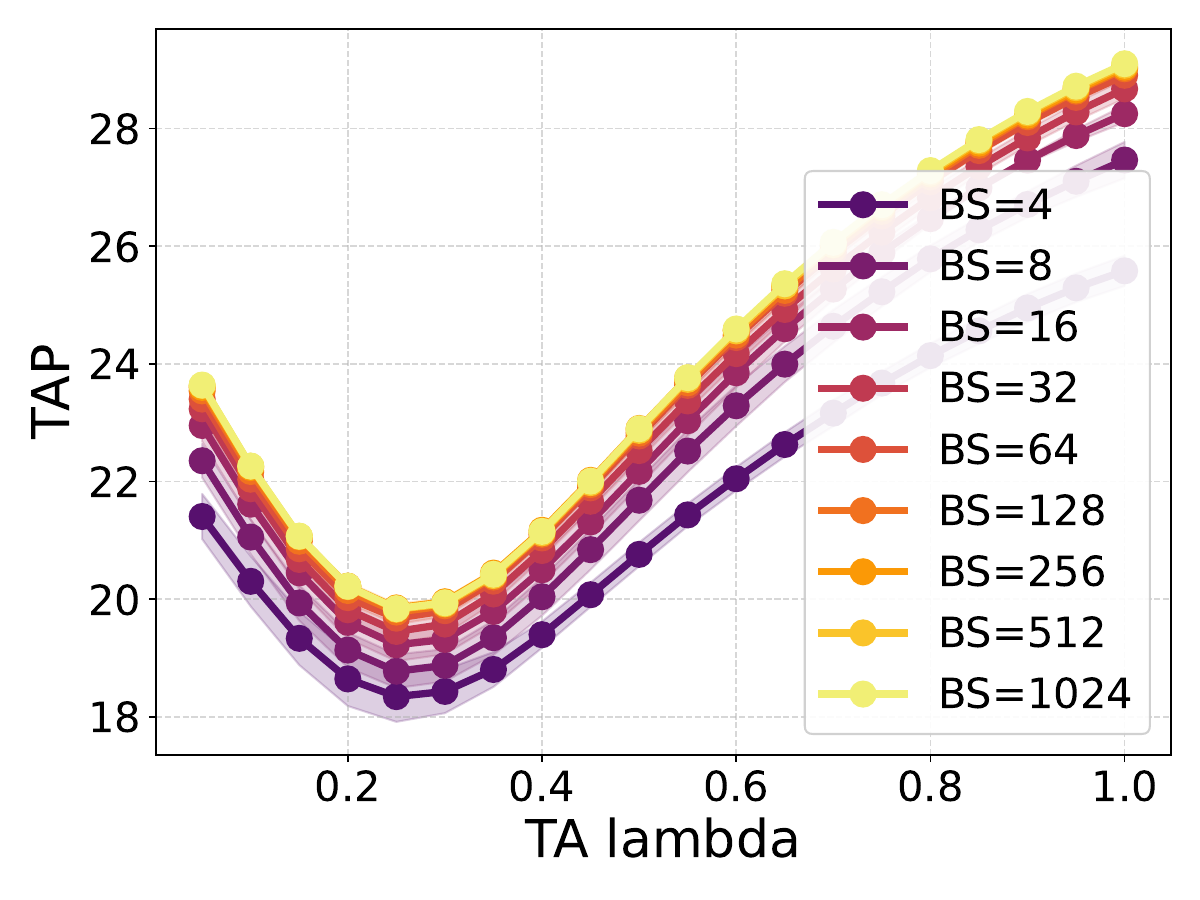}
    \end{minipage}
    \begin{minipage}{0.47\textwidth}
        \centering
        \includegraphics[width=\linewidth]{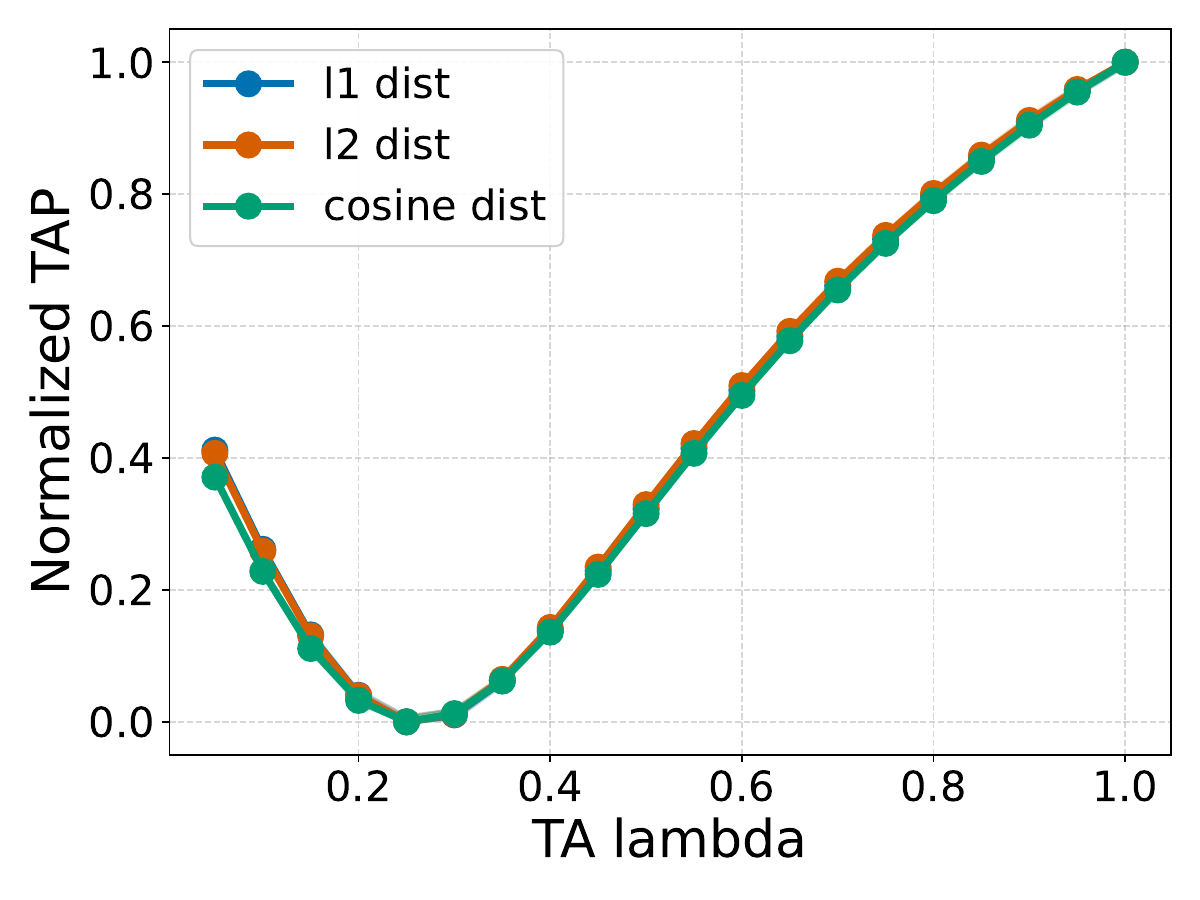}
    \end{minipage}
    \caption{
    \textbf{Left: \talign \vs Batch size.} When increasing the number of samples per task to compute \talign, the score becomes more stable. However, the selected model remains the same even when computing \talign  on very few images.
    \textbf{Right: \talign with different distance metrics.} \talign  is very robust to the distance metric used to compare fine-tuned and merged model features.
    }
    \label{fig:tas_bs_ablation}
\end{figure}

%% file: tex/06_analysis.tex
\section{Analyses}

\subsection{\Taskalign ablations} 
\label{sec:tas_ablations}

We ablate the design choices of \talign, in particular, 
the distance to compare features and the number of samples used to compute \talign. 
In \cref{fig:tas_bs_ablation} (left) we plot the \talign across different hyperparameter values when computed with different number of images. We observe that \talign is very robust to the images being used (shaded area corresponding to standard deviation across seeds is small). Moreover, it is highly data efficient, resulting in a significant reduction in computation. 
This also makes \talign  especially suitable when per-task data is scarce.
In \cref{fig:tas_bs_ablation} (right) we plot \talign across different hyperparameter values when measuring feature distance with $\ell_1,\ \ell_2$ or cosine distance (\ie $1~-$ cosine similarity). 
We see \talign is very robust to the choice of distance between features.

\input{tables/dinov2_merging}

\subsection{Can merging replace distillation?}
\label{sec:dinov2_merging}
In \cref{exp:dune_merging} we show that, in combination with \talign, 
model merging can be efficiently applied to a set of heterogeneous tasks.
However, 
DUNE was obtained by distilling already strong teachers in the respective downstream tasks. If such teachers were not available, can model merging be competitive with distillation? In this section we introduce an analysis experiment to address this question by starting from a strong generalist model, \ie DINOv2, and merging the corresponding fine-tuned models on the target tasks. The fine-tuning, merging methods and evaluation protocols are the same as used in \cref{tab:dune}.
However, instead of having DUNE as the base model, in \cref{tab:dinov2} we use DINOv2.

\input{floats/tv_analysis}

Since all fine-tuned models are initialized from DINOv2, most merged models perform well on ADE20k and NYUd, \ie tasks where DINOv2 is already strong. Several methods significantly outperform pretrained DINOv2 and even the distilled DUNE model. However, while some merged models surpass DUNE in normalized (average) performance, a trade-off exists between tasks (especially \bedlam and \mapfree), and no single model surpasses DUNE on all of them. We hypothesize that human mesh recovery and 3D scene reconstruction are further out-of-domain for DINOv2 than for DUNE, which was distilled on expert teachers for those tasks. These results are highly encouraging for future model merging using generalist backbones like DINOv2.

\subsection{Task vector norm analysis}
\label{sec:task_vector_norm_analysis}

To understand the differences between the standard CLIP benchmark and other studied settings (\ie LiDAR and DUNE), we analyze the task vectors directly.

\paragraph{Task vector imbalance.} 
\cref{fig:tv_analysis} (top) shows highly imbalanced task vector norms in the LiDAR and heterogeneous settings (DUNE/DINOv2 encoders), likely due to diverse tasks and fine-tuning protocols. Conversely, CLIP benchmark task vectors, which use a fixed fine-tuning protocol, have comparable magnitudes and significantly lower norms (see~\cref{app:tv_norm_comparison}). \cref{fig:tv_analysis} (bottom) shows that this norm imbalance biases the average task vector toward higher-norm tasks, creating an asymmetry which we hypothesize may hinder model merging.

\paragraph{Task subset experiments.}
To further explore the impact of task vector norms, we merge task subsets from the DUNE benchmark using four combinations: \{\ade, \nyudNS\}, \{\ade, \bedlamNS\}, \{\nyud, \bedlamNS\}, and \{\ade, \nyud, \bedlamNS\}, excluding \mapfree due to its high evaluation cost. 

\input{floats/task_combinations}

Results are reported in \cref{fig:task_combinations}.
All methods achieve their best normalized performance when merging \{\ade, \nyudNS\}, the tasks with the smallest and most similar norms. Performance consistently degrades when \bedlam is included, signaling increased ``merging difficulty''. Adding \bedlam also completely reverses the relative ranking of methods, highlighting that some approaches may be more sensitive to large and unbalanced norms.

\paragraph{Normalizing task vectors.}
If all tasks are to be weighted equally, a natural baseline is to normalize all task vectors before merging. We call this baseline \textbf{NormAvg}, where each task vector is rescaled to the smallest norm before averaging. This will naturally produce a more balanced norm and cosine similarity across tasks. However, in \cref{app:normavg} we show this does not lead to a promising performance. Thus, although a baseline, naively scaling task vectors does not seem to be an effective strategy. 

%% file: tables/dinov2_merging.tex
\begin{table}[t]
\caption{
\textbf{Teacher distillation \vs fine-tuning + merging.} We report results when merging models, initialized from DINOv2 (one of DUNE teachers), and fine-tuned on the DUNE benchmark. We also compare with the DUNE distilled model as a baseline.}

\label{tab:dinov2}
\centering
\begin{tabular*}{\textwidth}{@{\extracolsep{\fill}}lrrrrr}
\toprule
\makecell{ \\ } & \makecell{\ade \\ mIoU (\(\uparrow\))} & \makecell{\nyud \\ rmse (\(\downarrow\))} & \makecell{\bedlam \\ pa-pve (\(\downarrow\))} & \makecell{\mapfree \\ AUC (\(\uparrow\))} & \makecell{Normalized \\ Performance} \\
\midrule
DUNE$^\dagger$ & {45.6} & 0.330 & 62.5 & 94.7 & {0.847}\\ 
DINOv2 (Pretrained) & {48.3} & 0.303 & 63.3 & 92.1 & {0.871} \\
\midrule
TA$_{\text{w/\talign}}$ & {50.2} & 0.303 & 62.0 & 94.5 & {0.889}\\
STAR$_{\text{w/\talign}}$ & {46.8} & 0.310 & 65.0 & \textbf{95.5} & {0.863} \\
TIES$_{\text{w/\talign}}$ & {49.3} & 0.298 & 62.2 & 94.1 & {0.887}\\ 
Consensus$_{\text{w/\talign}}$ & {49.9} & 0.309 & \textbf{57.4} & 93.1 & {0.892} \\
TSV$_{\text{w/\talign}}$ & {50.3} & 0.296 & 63.4 & 94.8 & {0.893} \\
Iso-CTS$_{\text{w/\talign}}$ & 50.9 & \textbf{0.276} & 61.5 & 93.3 & \textbf{{0.915}} \\
AdaMerging$_{\text{w/\talign}}$ & {\textbf{52.8}} & 0.294 & 60.9 & 93.4 & {{0.909}}\\
\bottomrule
\end{tabular*}
\vspace{-0.1cm}
\end{table}

%% file: floats/tv_analysis.tex

\begin{figure}[t]
  \centering
  \setlength{\tabcolsep}{3pt}  
  \renewcommand{\arraystretch}{0} 

  \begin{tabular}{ccc}
    \includegraphics[width=0.32\linewidth]{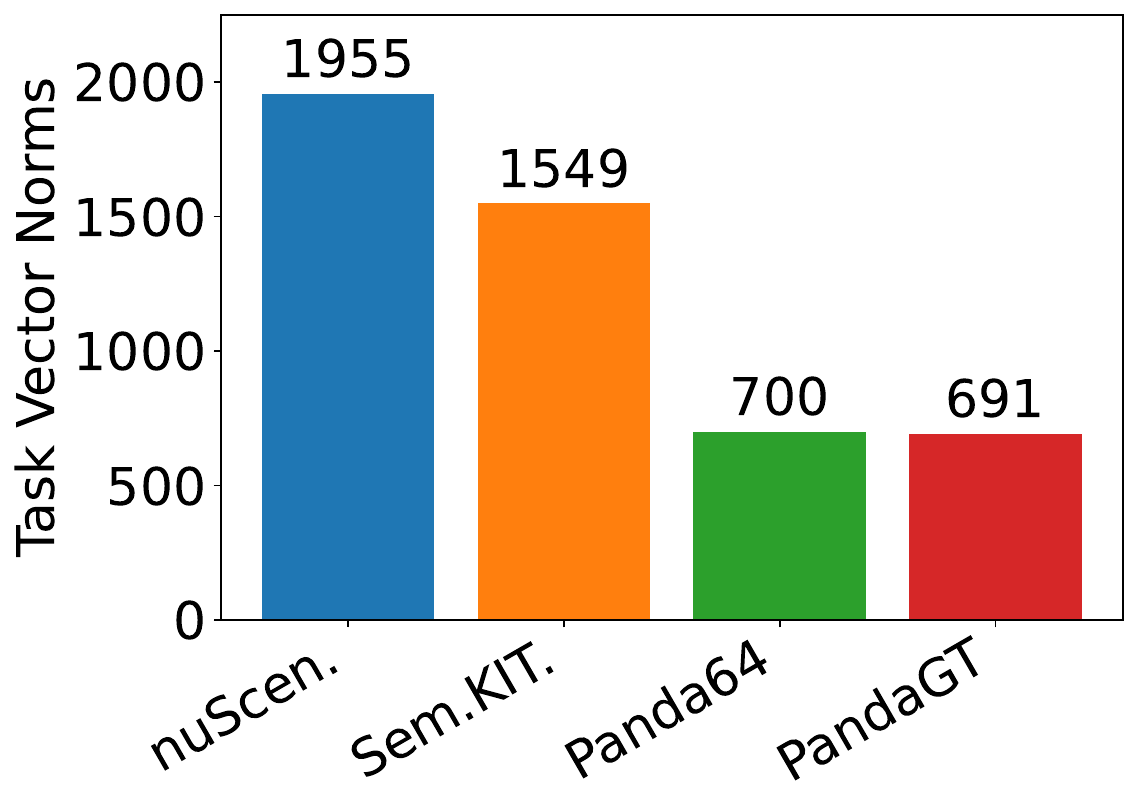} &
    \includegraphics[width=0.32\linewidth]{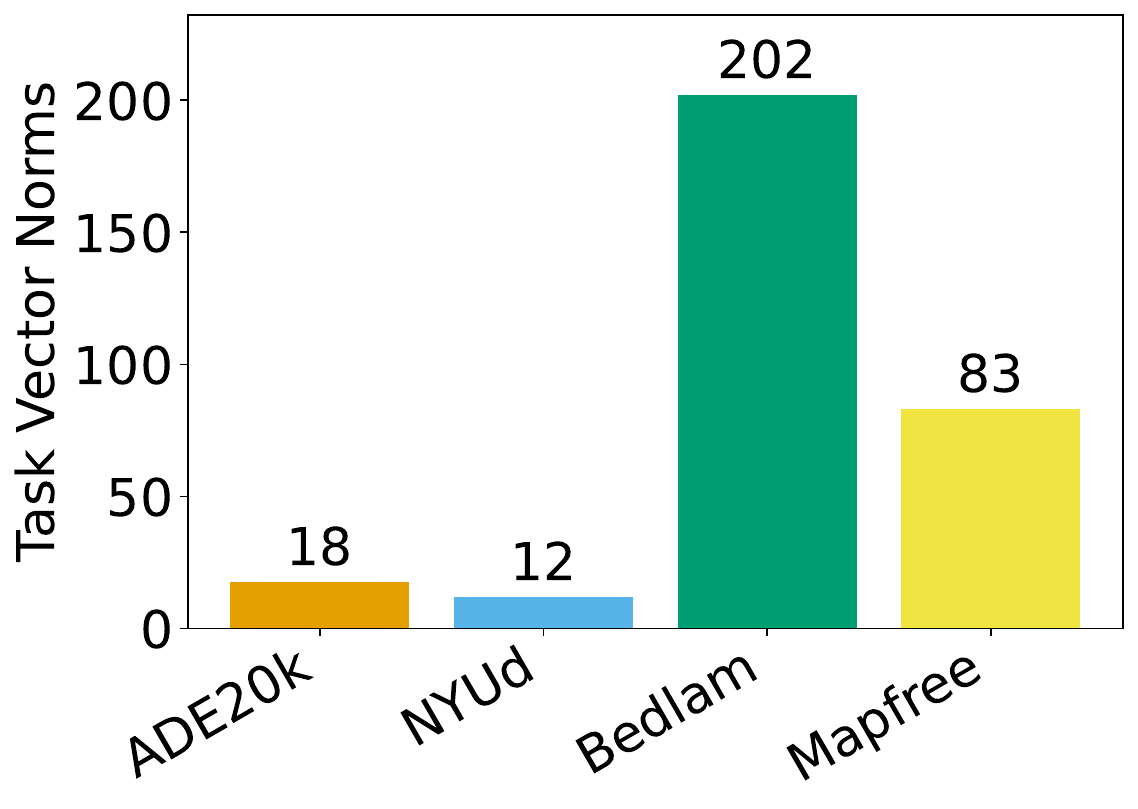} &
    \includegraphics[width=0.32\linewidth]{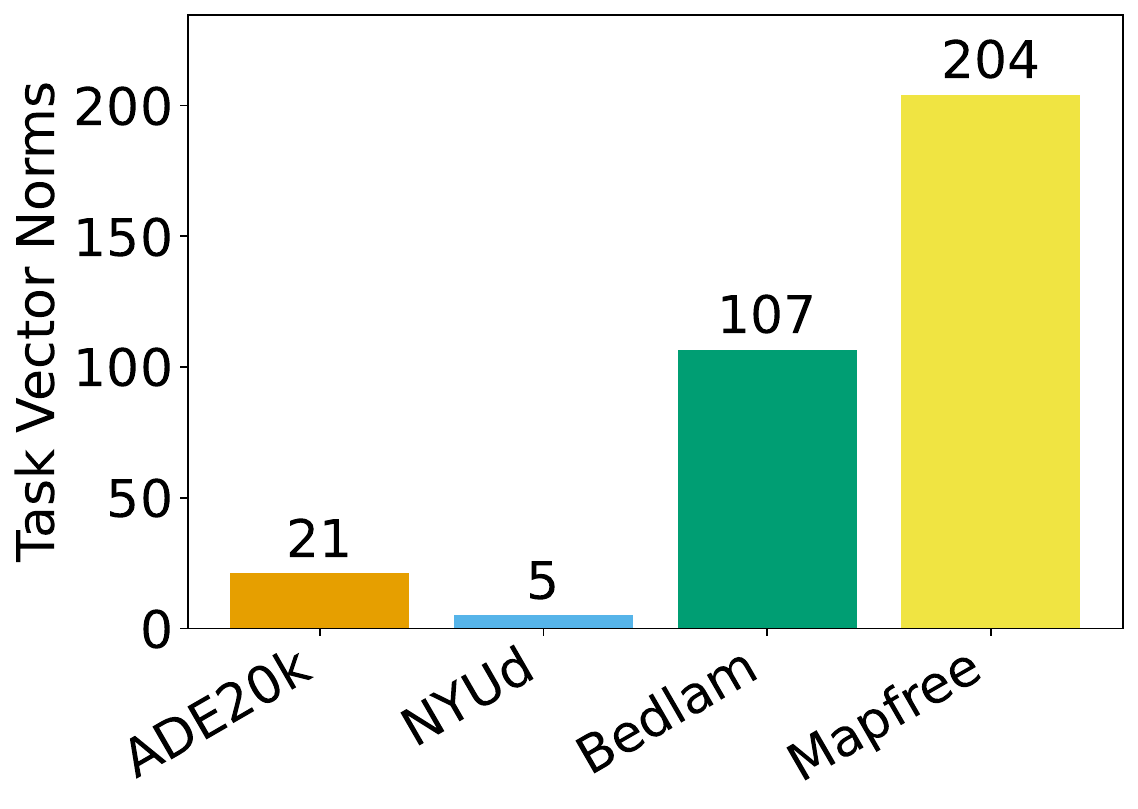}
     \\[4pt]
    \includegraphics[width=0.32\linewidth]{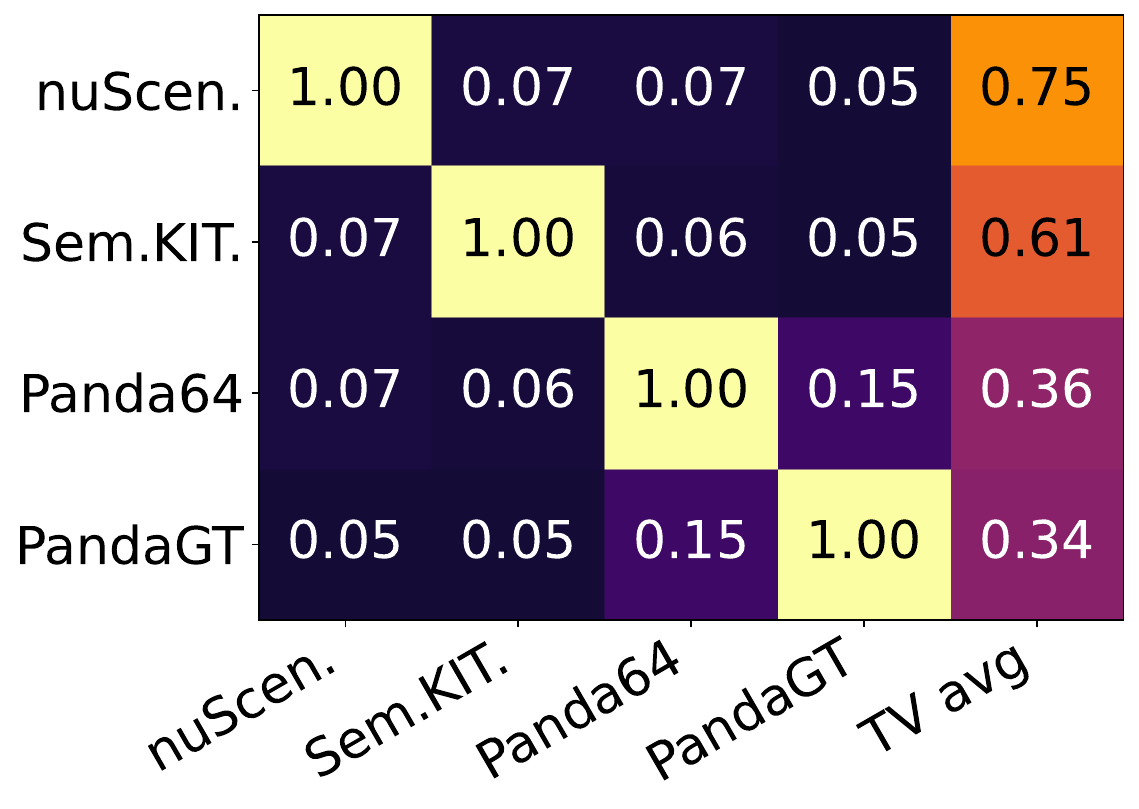} &
    \includegraphics[width=0.32\linewidth]{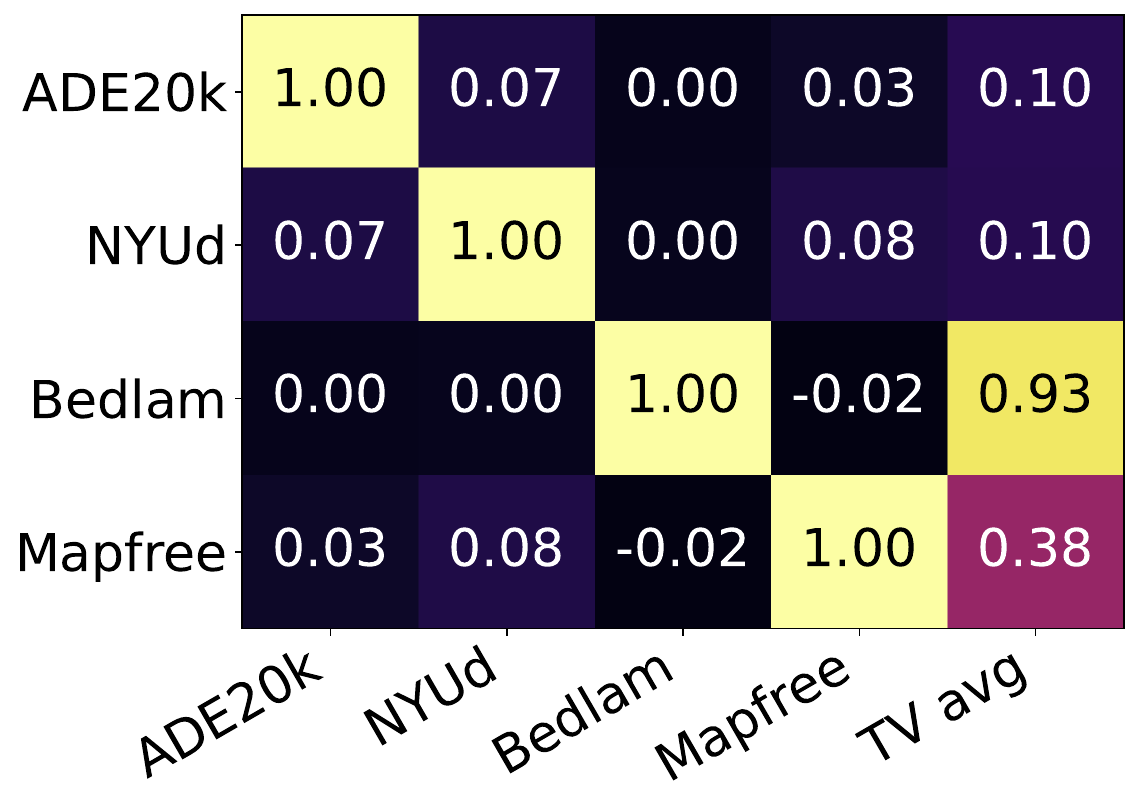} &
    \includegraphics[width=0.32\linewidth]{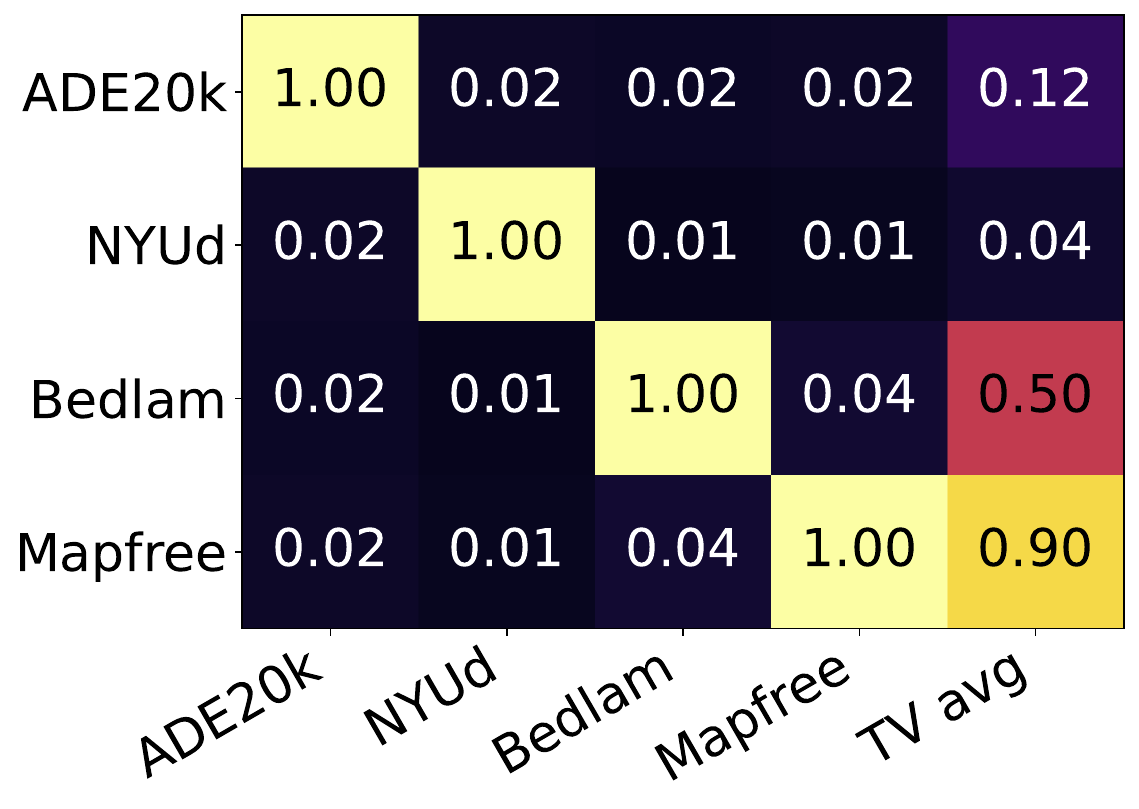}
  \end{tabular}
  \caption{
      \textbf{Top row:} Task Vector Norms for the LiDAR benchmark (left) and for the DUNE benchmark when finetuning models from DUNE (middle) or DINOv2 (right). \textbf{Bottom row:} Cosine similarity between task vectors and the average task vector. We observe that for all settings, task vector norms are rather large and unbalanced, strongly biasing the average task vector towards the task with high norm.
  }
  \label{fig:tv_analysis}
  \vspace{-10pt}
\end{figure}

%% file: floats/task_combinations.tex
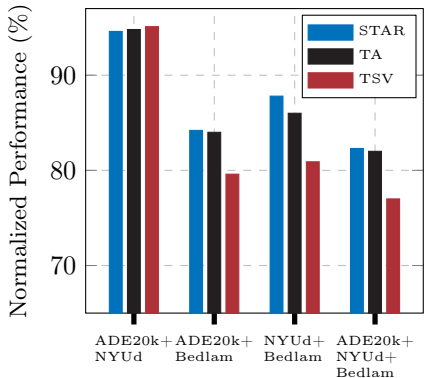
\begin{wrapfigure}[20]{r}{0.45\textwidth}
  \centering
  \vspace{-13pt}
  \begin{tikzpicture}
    \begin{axis}[
      scale only axis,
      width=0.37\textwidth,
      height=4cm,
      ybar,
      ymin=65, ymax=97,
      ylabel={\small Normalized Performance (\%)},
      xtick={0,1,2,3},
      xticklabels={
        {\makecell[l]{ADE20k+\\NYUd}},
        {\makecell[l]{ADE20k+\\Bedlam}},
        {\makecell[l]{NYUd+\\Bedlam}},
        {\makecell[l]{ADE20k+\\NYUd+\\Bedlam}}
      },
      xticklabel style={rotate=0, xshift=0pt, font=\tiny},
      xtick style={line width=2pt, color=black},
      xtick pos=bottom,
      bar width=0.16,            
      enlarge x limits=0.2,     
      grid=major,
      major grid style={dashed,gray!50},
      area legend,
      legend style={
        at={(0.98,0.98)},        
        anchor=north east,       
        inner sep=2pt,           
        font=\tiny,
        fill=white, fill opacity=1, draw opacity=1,text opacity=1,
        legend cell align=left   
      }
    ]

    \addplot+[draw=RoyalBlue, fill=RoyalBlue]   
    coordinates { (0,94.6) (1,84.2) (2,87.8) (3,82.3) };
    \addplot+[draw=Black, fill=Black]    
    coordinates { (0,94.8) (1,84.0) (2,86.0) (3,82.0) };
    \addplot+[draw=Maroon, fill=Maroon]  
    coordinates { (0,95.1) (1,79.6) (2,80.9) (3,77.0) };

    \legend{STAR, TA,
    TSV}
    \end{axis}
  \end{tikzpicture}
  \caption{
  \textbf{Performance \vs task subsets.} When task vector norms are small and balanced (left-most), merging methods perform significantly better than when some task vectors are much larger than others.
  }
  \label{fig:task_combinations}
\end{wrapfigure}

%% file: tex/07_conclusion.tex
\section{Conclusion}

In this paper, we explore model merging in vision beyond the widespread CLIP-based classification benchmark. 
Because most vision tasks rely on task-specific trainable decoders, downstream performance-based hyperparameter selection becomes impractical. To address this, we propose \textit{Task Alignment}, a simple proxy of downstream performance that selects merging hyperparameters at a fraction of the cost. On top of being very cost efficient, it requires no access to task-decoders or labels, making it widely applicable: 
\textit{any} merging method can use it for \textit{any} set of tasks,
and it can even serve as a loss function to generalize some learning-based methods like \adamerge.
Our extensive experiments
show that \talign-based hyperparameter selection leads to 
near-optimal
model selection on a diverse set of model merging benchmarks, including 
CLIP classification. 
Moreover, task vectors in our newly introduced merging settings are significantly larger and more unbalanced than in the CLIP setting, 
testing the limits of existing merging methods. Explicitly addressing large, unbalanced task vectors in model merging is a promising direction for future work, 
and our preliminary findings provide initial guidance toward this goal.

%% file: tex/08_appendix.tex
\appendix

\renewcommand{\thesection}{\Alph{section}}
\gdef\theHsection{\Alph{section}}


{
\setlength{\textfloatsep}{8pt plus 2.0pt minus 2.0pt}
\setlength{\floatsep}{10pt plus 2.0pt minus 2.0pt}

\setcounter{page}{1}

\begin{center}
    \Large --- Supplementary Material ---  \\
\end{center}

\setcounter{tocdepth}{2}
\startcontents
\printcontents{}{1}{\section*{Table of contents}}



\section{Additional details on the experimental protocol}
\label{app:experimental_details}

In \Cref{sec:protocol} we provide an overview of the different evaluation protocols used in our experiments. In this section we provide additional details.

\subsection{CLIP-based classification} 

We strictly follow the implementation of the evaluation protocol released by \cite{gargiulo2025task} using their public code\footnote{ \url{https://github.com/AntoAndGar/task_singular_vectors/tree/main}}. For convenience, we quote the experimental details provided in \cite{gargiulo2025task}:

``We evaluate our approaches over three different suites of
tasks having cardinality 8, 14, and 20, respectively. The
first one, introduced in \cite{ilharco2023editing}, consists of datasets: Cars \cite{krause2013cars}, DTD \cite{cimpoi2014dtd}, EuroSAT \cite{helber2019eurosat}, GTSRB \cite{stallkamp2011gtsrb}, MNIST \cite{lecun1998mnist},
RESISC45 \cite{cheng2017resisc45}, SUN397 \cite{xiao2016sun397}, and SVHN \cite{netzer2011svhn}. 
The benchmark with 14 tasks builds on the preceding one, incorporating six additional datasets: 
CIFAR100 \cite{krizhevsky2009cifar}, STL10 \cite{coates2011stl10},
Flowers102 \cite{nilsback2008flowers102}, OxfordIIITPet \cite{parkhi2012oxfordpets}, PCAM \cite{veeling2018pcam},
and FER2013 \cite{goodfellow2013fer2013}. 
Finally, the 20-tasks benchmark includes the preceding 14 plus the following six: 
EMNIST \cite{cohen2017emnist}, CIFAR10 \cite{krizhevsky2009cifar}, Food101 \cite{bossard2014food101}, 
FashionMNIST \cite{xiao2017fashionmnist}, RenderedSST2 \cite{socher2013sst}, and KMNIST \cite{clanuwat2018kmnist}. We evaluate our method on three variants of the CLIP \cite{radford2021clip}
model, each employing a different size of ViT \cite{dosovitskiy2021an} visual encoder: 
ViT-B-32, ViT-B-16, and ViT-L-14.
The main benchmarks involve merging 8, 14, and 20 tasks,
mirroring the experimental setup described in Wang et al.
\cite{wang2024localizing}.''

\subsection{LiDAR tasks}

To the best of our knowledge, this is the first application of model merging to point cloud encoders for LiDAR datasets. We therefore detail the protocol that we propose.

\paragraph{Datasets.}
The characteristics of the LiDAR sensors and the specifications of the datasets used can be found in~\cref{tab:lidar_datasets}. Notably, for all four datasets a different LiDAR sensor has been used for acquisition. The reported performances are calculated for nuScenes and SemanticKITTI on the official train/validation splits. For Panda64 and PandaGT we adopt the data splits and class definitions from~\cite{puy2024three}. While nuScenes, SemanticKITTI and Panda64 are obtained by LiDAR sensors with $360^\circ$ coverage, PandaGT is based on a forward-facing sensor only.

\input{tables/appendix/lidar_datasets}

\paragraph{Pretraining.} 
We utilize a point cloud backbone pretrained via ScaLR~\cite{puy2024three}, a multi-modal image-to-LiDAR distillation method. ScaLR leverages patch-to-point correspondences to distill a Vision Foundation Model (VFM) (concretely DINOv2~\cite{oquab2024dinov2}) into a WaffleIron~\cite{puy23waffleiron} encoder. During pretraining, a linear projector maps the backbone output to the feature space of the VFM. The model is pretrained on the nuScenes, SemanticKITTI, Panda64 and PandaGT dataset. Code and checkpoints are available at the official repository\footnote{\url{https://github.com/valeoai/ScaLR}}.

\paragraph{Fine-tuning.} As fine-tuned models for each dataset, we use the provided checkpoints in~\cite{puy2024three} as they are the highest performing models publicly available at the time of writing. For the fine-tuning in~\cite{puy2024three}, the linear projector layer used during pretraining is removed and replaced by a linear classification layer. The entire backbone, together with this classification layer, is then trained in a supervised manner on the respective dataset. 

\paragraph{Model merging.}
We adhere to the model merging algorithms from the literature, adapting them only where necessary. SVD-based approaches, such as STAR and TSV, were originally designed for two-dimensional weight matrices, e.g. in attention modules. Since WaffleIron lacks attention modules, we apply SVD to the MLP layer matrices, which account for $96.3\%$ of the model parameters. Applying the Consensus~\cite{wang2024localizing} method to the WaffleIron encoders we noticed a numerical instability which prevents the necessary training of the decoder. In a closer investigation, it appeared that the task vector compression introduced in~\cite{wang2024localizing} is responsible for the instability. Disabling the task vector compression for the embedding layer, as well as the layer normalization layers ensured numerical stability while affecting only $3.2\%$ of the parameters. Furthermore, we also trained and evaluated the linear decoder for this model merging method with the FP32 precision instead of FP16.

\paragraph{Linear decoder training.} We train a task-specific linear decoder on top of the merged, frozen encoder. Following the optimization parameters for linear probing in~\cite{puy2024three}, we found that 5~epochs for nuScenes and SemanticKITTI, and 10~epochs for Panda64 and PandaGT, are sufficient for convergence. To reduce computational cost, we therefore restrict training to these numbers of epochs. Training for nuScenes takes approximately 13.5~hours on a single NVIDIA V100 (32GB VRAM); SemanticKITTI, Panda64, and PandaGT require 9.5, 3.7, and 3.7~hours on a single NVIDIA A100 (80GB VRAM).

\paragraph{\talign calculation.} We align our evaluation with the protocol established for image models. Since reliable \talign scores can be calculated from a small subset of data, we use a fixed number of 128~scenes. For each scene, we randomly sample 1369 points (equivalent to the $37\times37$ patches of an image). We concatenate the per-point features of sampled points to obtain a single vector per scene. The \talign score is calculated on the $128\times(1369\cdot 768)$ matrix, where 768 is the WaffleIron feature dimension.

\input{floats/cost_vs_perf_lidar}

\paragraph{Compute resources for hyperparameter selection.}
If a training of the decoder is necessary, 30~gpu~hours \textbf{per hyperparameter} are needed. This has to be contrasted to a selection time of only 7~minutes with \talign per hyperparameter.}
In~\cref{fig:cost_vs_perf_lidar} we plot the performance \vs the hyperparameter selection cost, highlighting that the \talign-based selection is reducing the computational cost in the LiDAR setting by 2 orders of magnitude.

\subsection{Heterogeneous vision tasks}

\paragraph{\Finetuning settings and costs.} We follow two different setups when \finetuning decoders and evaluating models: the \textit{ablation protocol}, where we limit the computational budget per \finetuning; and the \textit{long \finetuning protocol}. For the long \finetuning protocol, we follow the same settings as described in \cite{sariyildiz2025dune}:
\begin{itemize}
    \item \ade: We finetune a linear head as in \cite{oquab2024dinov2} for 80k iterations.
    \item \nyud: Similarly, we follow \cite{oquab2024dinov2}, 
    and finetune for 48k iterations.
    \item \bedlam: We follow \cite{multihmr} and finetune for 200k iterations.
    \item \mapfree: We follow \cite{mast3r} and finetune for 10 epochs on several datasets with 3D correspondences.
\end{itemize}

Of the above, \mapfree \finetuning is by far the most expensive as it involves the largest decoder. 
For this protocol, \mapfree \finetuning requires 4 H100s and takes around 4 days to complete. Other tasks take less than 24 hours on a single H100. 
For the shorter \textit{ablation protocol} we do as follows:
\begin{itemize}
    \item \ade: We finetune a linear head as in \cite{oquab2024dinov2} for 40k iterations.
    \item \nyud: Similarly, we follow \cite{oquab2024dinov2}, 
    and finetune for 24k iterations.
    \item \bedlam: We follow \cite{multihmr} and finetune for 60k iterations.
    \item \mapfree: We follow \cite{mast3r} and finetune for 3 epochs only on the \mapfree dataset (which was part of the training dataset mix used in \cite{mast3r}).
\end{itemize}
With this ablation protocol, the \mapfree \finetuning can be done in about 5 hours with 4 H100s and the other tasks take less than 12 hours with a single GPU.

\paragraph{Keeping distillation projectors.} Interestingly, in the original work of \cite{sariyildiz2025dune}, authors observe that for \ade, keeping the distillation projector for the DINOv2 teacher (\ie, a small teacher-specific head that projects student encoder features 
into the teacher feature space) 
improves performance significantly. We find that keeping the distillation projector improves performance, \textit{even if the teacher projector is frozen}, \ie, not part of the per-task \finetuning or merging. As in \cite{sariyildiz2025dune}, for other tasks, we do not see improvements when keeping the teacher projectors. Note that when we perform merging starting from DINOv2, which was not distilled as DUNE, there are no projectors to be used.

\subsection{Baseline with normalized task vectors} 
\label{app:normavg}

We refer to the baseline where each task vector is rescaled to the smallest norm before averaging as \textbf{NormAvg}.
This method, discussed in \cref{sec:task_vector_norm_analysis}, is implemented as follows:
\begin{equation}
    \theta^l_{\text{Merged}} = \theta^l_0 + \sum_{t=1}^T \lambda^l_t \tau^l_t, \ \text{where}\ \  \lambda^l_t = \dfrac{\min_j ||\tau^l_j ||_2}{|| \tau^l_t ||_2}.
\end{equation}
That is, for each layer $l$, we scale the task vector $\tau^l_t$ so that it has norm equal to the smallest task vector. Recall $\tau^l_t := \theta^l_t - \theta^l_0$. We also tested normalizing the full task vectors (\ie a single $\lambda_t$ per task) but found per-layer normalization led to slightly better results.

Under the ablation protocol of the DUNE benchmark, per-layer NormAvg achieves a normalized performance of 0.878. This performance is marginally better than simple model averaging (0.877) but worse than TA (0.886).

\subsection{Hyperparameter search space}

In \cref{fig:talign_scatters} we plot the correlation between TAP and performance for different hyperparameters ablated. In \cref{app:tab:clip_hyp_search}, \cref{app:tab:lidar_hyp_search} and \cref{app:tab:het_setting_hyp_search} we report the hyperparameters used for the CLIP, the LiDAR and the Heterogeneous settings, respectively.

Note that in \cref{sec:protocol} we report two different protocols for the heterogeneous setting due to the high cost of training some of its decoders. For the final numbers reported in \cref{tab:dune} and \cref{tab:dinov2}, we extend/fine-grain the hyperparameter search since we only compute the efficient TAP for selection. The selected hyperparameters for each method are reported in \cref{app:tab:selected_long_eval}. For the LiDAR setting we also ablate the Common Space Fraction (csf) for the Iso-CTS method and find the optimal hyperparameters to be $\alpha=0.8$ and csf~$=1.0$, which is actually equivalent to Iso-C~\cite{marczak2025notaskleftbehind}.

\input{tables/appendix/clip_hyp_search}

\input{tables/appendix/lidar_hyp_search}

\input{tables/appendix/het_setting_hyp_search}

\input{tables/appendix/selected_hyps_long_eval_het_setting}


\section{Further discussion on related work}
\label{app:related_work_extended}

\input{tex/extended_rel_work_v1}

\section{Further experiments on the CLIP setting}
\label{app:further_exps_clip}

In \cref{tab:clip_summary} of the main paper we present the results comparing \talign-based hyperparameter selection and evaluation-based hyperparameter selection for ViT-B-32 and ViT-L-14 architectures, where we see that the drop in performance when using \talign is minimal despite not using any labels and only a few images per dataset. While not shown in \cref{tab:clip_summary} due to space constraints, we also performed our experiments on a ViT-B-16 architecture, following the evaluation protocol in~\cite{gargiulo2025task}. In \cref{tab:clip_full} we show the full results on the three architectures and observe a similar behavior with the ViT-B-16 architecture as with the two others.

\input{tables/appendix/clip_full}

\section{Further experiments on the LiDAR setting}
\label{app:further_exps_lidar}

\input{tables/appendix/lidar_table_full}

\subsection{TAP vs. Evaluation-based Hyp. selection} In~\cref{tab:app:lidar_full}, we report the performance of various model merging algorithms using both compute-intensive evaluation-based ($_\text{w/\HS}$) and our \talign-based ($_\text{w/\talign}$) selection. As visualized in \cref{fig:talign_scatters}, the hyperparameters selected by \talign provide close-to-optimal results; specifically, the average mIoU remains within $0.8$ percentage points of the optimal value across all methods. Furthermore, while the original AdaMerging is incompatible with non-categorical tasks, our \talign-adapted version provides highly competitive results. 


\subsection{Task subset experiments}
As reported in~\cref{fig:tv_analysis}, we observe highly imbalanced task vector norms for the LiDAR setting, which motivated our task subset experiment described in \cref{sec:task_vector_norm_analysis}. In \cref{fig:task_combinations} we conducted the experiment on the Heterogeneous tasks setting; for completion, here we conduct a similar study on the LiDAR setting. We select nuScenes, as it is the task with the highest task vector norm and Panda64 and PandaGT, which have the lowest. The results are shown in~\cref{fig:app:task_combinations_lidar}. We reach a similar conclusion: \textit{all merging methods perform best when task vectors norms are smaller and more similar}.

\subsection{Merging synergy on PandaGT}
Interestingly, we observe that several merged models achieve better performance than the fine-tuned models for the PandaGT dataset. See, for instance, AdaMerging in \cref{tab:app:lidar_full}. This also results in a normalized performance above one when merging the Panda64 and PandaGT datasets in \cref{fig:app:task_combinations_lidar}. Indeed, we already see a synergy between the fine-tuned models, before any merging. While the pretrained model achieves an mIoU of 35.4 on PandaGT, the model fine-tuned on Panda64 achieves 39.2, nuScenes 36.2 and SemanticKITTI 36.7, all above the performance of the pretrained model. This is somewhat surprising as, in general, when fine-tuning a model on a specific dataset one expects performance on other datasets to decrease slightly. 
%
\input{floats/task_combinations_lidar}

\section{Further experiments on the heterogeneous setting}

In the main paper we show that \talign and performance are strongly correlated and that \talign-based hyperparameter selection is very close to optimal. In particular, \cref{fig:talign_scatters} depicts the aforementioned results for the CLIP, LiDAR and heterogeneous tasks settings. Note, that by default, our experiments on the heterogeneous task setting use DUNE as the base model, however in \cref{sec:dinov2_merging} we also conduct an analysis where we show that directly finetuning a generalist encoder (i.e. DINOv2) on the same heterogeneous tasks and then merging them back leads to very competitive results compared to the expensive distillation process from ``expert'' teachers followed in DUNE~\cite{sariyildiz2025dune}. For completeness, in this section we also present the same scatter plots between normalized performance and \talign as presented in \cref{sec:dinov2_merging}, but this time using the DINOv2 as base model for the heterogeneous tasks setting.
In line with previous results, in \cref{fig:app:dinov2_ablation} we observe that \talign-based hyperparameter selection leads to close-to-optimal performance.

As described in \cref{sec:protocol} given the cost of decoder finetuning in the heterogeneous tasks settings, we introduce a shorter decoder training protocol when doing evaluation-based hyperparameter selection. Thus, note that the normalized performance reported in \cref{fig:app:dinov2_ablation} is not directly comparable with the performance reported in \cref{tab:dinov2} of the main paper, where we finetune the \talign-selected hyperparameters with the full protocol described in DUNE~\cite{sariyildiz2025dune} and evaluate them on the full test sets (not custom validation sets).

\input{floats/dinov2_ablation}

\section{Merging with frozen decoders}
\label{app:frozen_heads}

A baseline to alleviate the cost of hyperparameter selection 
is to keep decoders frozen when evaluating different hyperparameters. This way the costly decoder training for each hyperparameter setting can be skipped. However, even with frozen decoders, performing such evaluations requires access to a labeled validation set and, more importantly, requires the use of potentially expensive and complex decoders. 

On the other hand, \talign is \textit{decoder-free}. When merging a heterogeneous set of tasks, different task-specific systems may rely on different databases, codebases, python environments, etc; which are all needed
for downstream performance evaluation.
Instead, TAP reduces selection to a common encoder-level computation.
This simplicity is the point: the same criterion can be applied across CLIP classification, LiDAR segmentation, and heterogeneous 2D/3D vision, regardless of downstream evaluation code. 
Moreover, in the heterogeneous setting we observe that downstream evaluation using frozen heads is still $\sim 6\times$ more expensive than TAP, which requires only an encoder forward.

\begin{figure}[t!]
    \centering
    \includegraphics[width=0.9\linewidth]{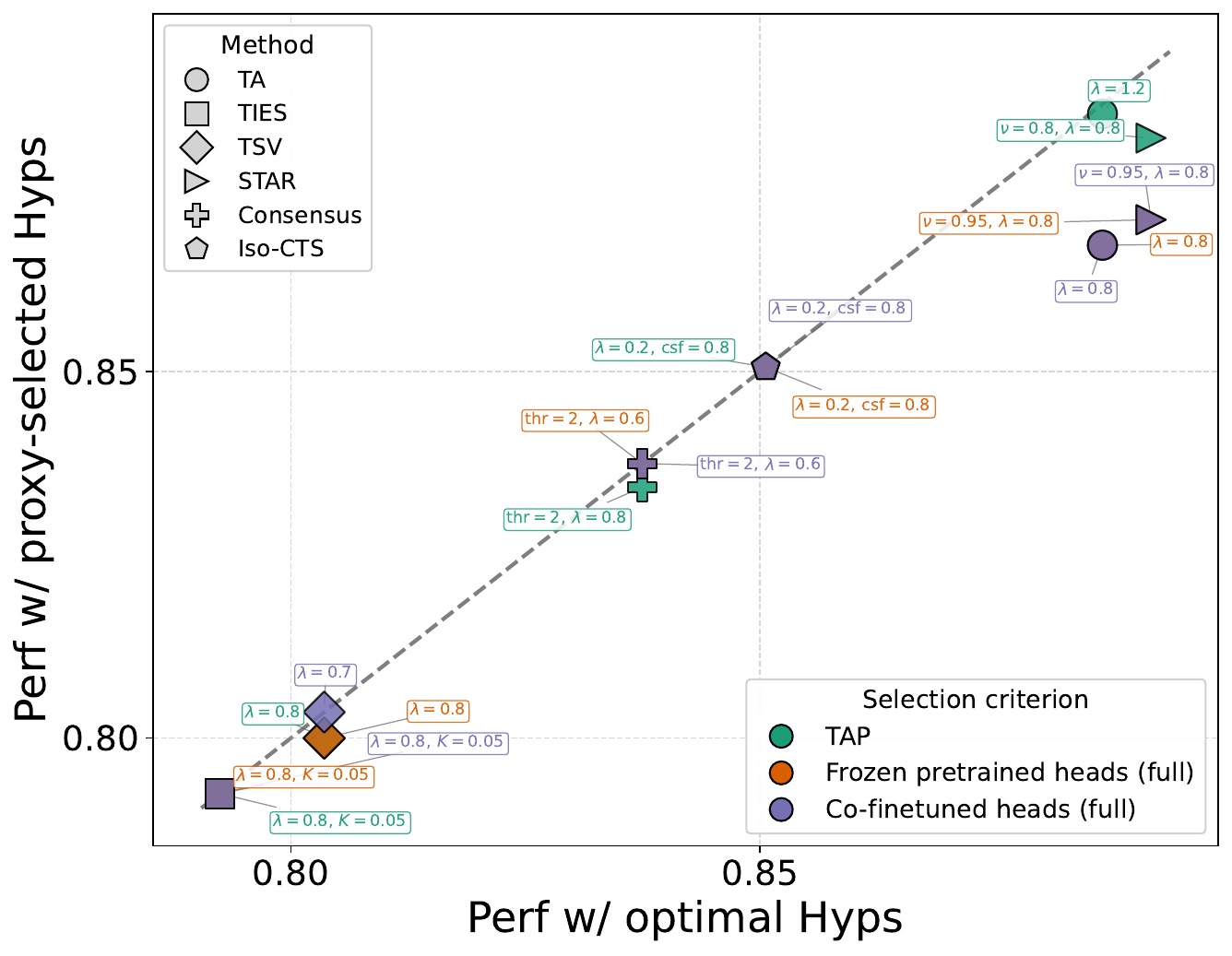}
    \caption{\textbf{Frozen heads (decoders) ablation}. We compare TAP-based hyperparameter selection with the frozen-heads baselines and observe that TAP achieves comparable or better performance than the frozen decoders while only relying on encoder features and no downstream labels.}
    \label{fig:frozen_heads}
\end{figure}

In \cref{fig:frozen_heads}, we compare TAP-based selection with two frozen decoders (heads) baselines. \textit{Frozen pretrained heads:} That is, keeping the frozen heads trained on top of the pretrained (base) model, \ie $\theta_0$ from \cref{sec:related}. \textit{Frozen co-finetuned heads:} That is, the heads co-finetuned with the encoders for the respective tasks. Note, the respective fine-tuned encoders are the $\theta_t$ from \cref{sec:related} that are then merged. We observe that TAP selection matches or surpasses evaluation-based selection with complex downstream decoders in the heterogeneous setting.

\section{Analyzing the fine-tuned models}
\label{app:tv_norm_comparison}

\begin{figure}[t!]
    \centering
    \includegraphics[width=0.99\linewidth]{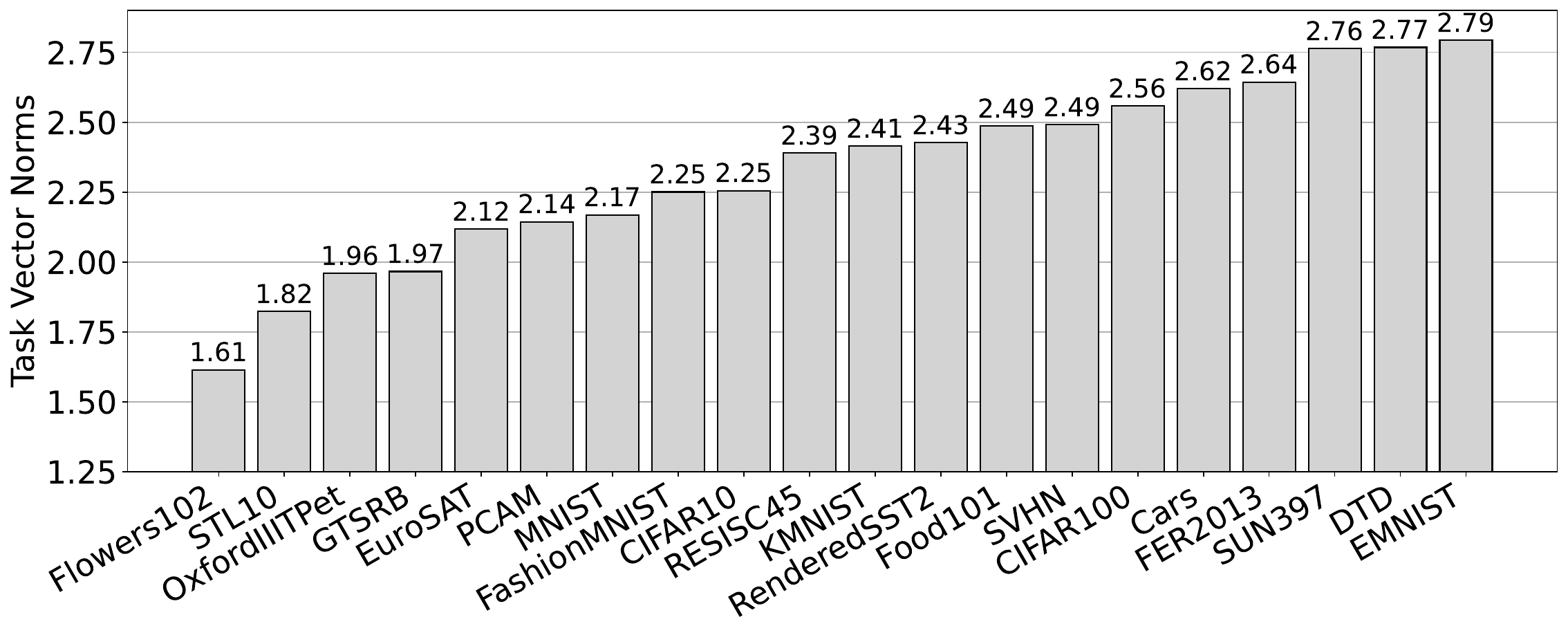}
    \caption{\textbf{Task vector norms for the 20-task CLIP benchmark.}
        For each task, we report the $\ell_2$ norm of the difference between its flattened parameter vector and that of the base model.}
    \label{fig:clip_benchmark_norms}
\end{figure}

\paragraph{Task vector norms for the CLIP and DUNE benchmarks.}
In~\cref{fig:clip_benchmark_norms}  we plot the task vector norm for the 20 tasks of the CLIP benchmark, using the weights released by TSV~\cite{gargiulo2025task}.\footnote{\url{https://github.com/AntoAndGar/task_singular_vectors}} 
Since all models are ViT-Base, the absolute norm values are comparable to those in~\cref{fig:tv_analysis} for the DUNE benchmark. We see that even in the ``furthest'' case (EMNIST) the TV norm is much smaller than the ``closest'' of the four DUNE tasks (2.79 vs 12).

\begin{figure}[t!]
    \centering
    \includegraphics[width=0.99\linewidth]{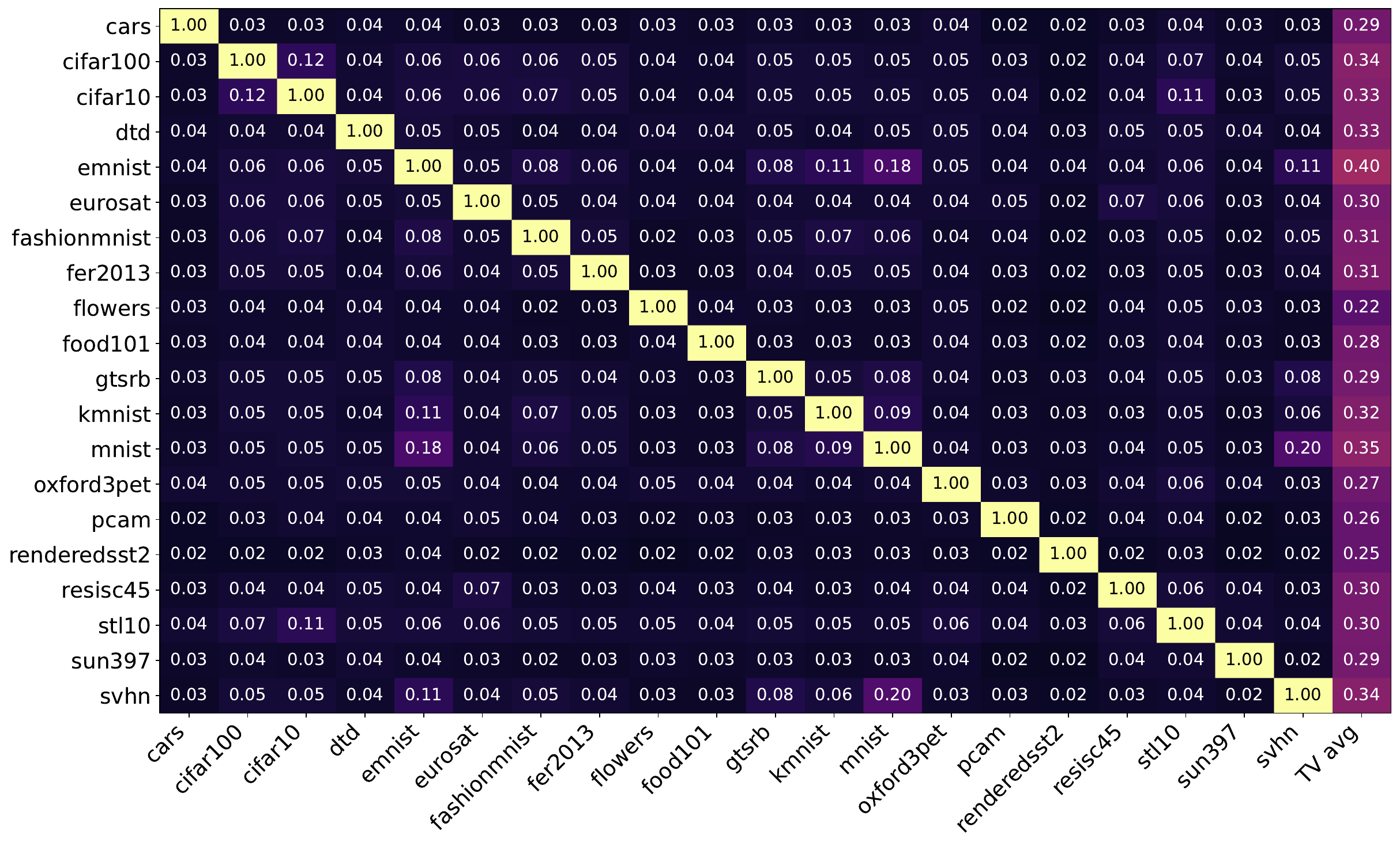}
    \caption{Cosine similarity between the task vectors of the CLIP merging setting and the average task vector. Given that all task vector norms are of similar magnitude, the average task vector is similarly aligned with all tasks.}
    \label{fig:clip_benchmark_cos_sim}
\end{figure}

\paragraph{Cosine similarity across task vectors.}
In \cref{fig:clip_benchmark_cos_sim} we show the cosine similarity across the task vectors of the CLIP tasks. While in \cref{sec:task_vector_norm_analysis} we saw that the large task vector norm imbalance introduced an asymmetry in the average task vector, naturally biased towards the largest norms, for CLIP we observe all tasks seem to be similarly aligned with the average task vector.

\paragraph{Task vectors across layers.}
We repeat the analysis from~\cref{fig:tv_analysis,fig:clip_benchmark_norms} when looking this time at the task vector norms per layer.
For each layer, we flatten and concatenate all parameters in that layer, then calculate the $\ell_2$ norm between the base and fine-tuned models. In~\cref{fig:tv_per_layer} we plot the task vector norms for each of the 12 layers of the ViT-Base model.
We show the norms for each of the 4 tasks of the DUNE benchmark, as well as the average over all 20 CLIP benchmark tasks (with 
standard deviation shown as error bars).

\begin{figure}[t!]
    \centering
    \input{fig/param_change_per_layer}
    \caption{\textbf{Task vector norms per layer for the CLIP and DUNE benchmarks.} For CLIP we plot the average and standard deviation over all 20 tasks. See~\Cref{tab:tv_per_layer} for the full set of numbers. }
    \label{fig:tv_per_layer}
\end{figure}


\section{\talign as a loss for \adamerge}
\label{app:ada}

Even though we present \talign as a proxy for final model performance, one can also look at the \talign objective from a differentiable optimization perspective and transform it into a \textit{differentiable loss function}.
In this section, we describe how to obtain such loss and how we have applied it to serve as the optimization objective in \adamerge.

As discussed in \cref{sec:related}, in \adamerge the $\lambda$ merging coefficients are computed through parameter optimization. The authors proposed to use entropy minimization on unlabeled target samples as the surrogate objective for optimizing the merging coefficients, therefore making the method naturally applicable to classification tasks.
In order to make it applicable to more generic tasks, we define a loss based on our general \talign structure.

\subsection{\Taskalign as a loss}
\label{app:ada_talign-loss}

Let $f(x;\theta_{merged})$ denote the encoder of the merged model, hereby referred to as the \emph{student}; and 
$f(x;\theta_t)$ the finetuned encoder of the task-$t$ as the \emph{teacher}. 
Given an unlabeled dataset $D_t$ for task $t$, let
$d : \mathcal{F} \times \mathcal{F} \to \mathbb{R}^+$ be a dissimilarity function defined over the feature vectors in the encoder space $\mathcal{F}$.  
We then define the \textit{per-task alignment loss} as a function of the merging coefficients $\lambda$ as:

\begin{equation}
\mathcal{L}_{\text{align}}^{(t)}(\lambda)
\;=\;
\frac{1}{|D_t|}\sum_{x \in D_t} 
d\!\left(f\!\left(x;\theta_{\mathrm{merged}}(\lambda)\right),\,
         f(x;\theta_t)\right),
\end{equation}

where $\theta_{\mathrm{merged}}(\lambda)$ is the merged encoder
defined in Eq.~\ref{eqn:th}, that in this case does not employ $\mu$ and $\phi$ is the identity.
To obtain a single set of coefficients across all tasks, we then minimize the average \taskalignloss:

\begin{equation}
\mathcal{L}_{\text{align}}(\lambda) 
\;=\; \frac{1}{T}\sum_{t=1}^{T} \mathcal{L}_{\text{align}}^{(t)}(\lambda).
\end{equation}

It is worth emphasizing the distinction between the original use of \talign\ as a 
performance proxy and its role here as a loss function. While the proxy does not need to be differentiable, using it as an optimization objective requires $d(\cdot,\cdot)$ to be almost everywhere differentiable, so that gradients can be computed during training.
We discuss this and other practical considerations below.

\subsection{Practical considerations and implementation details}

Although the use of this framework appears straightforward, its application requires careful attention to a few implementation details.

\paragraph{Choice of dissimilarity function.}
When using \talign as a proxy for performance, we observed little difference between setting $d(\cdot,\cdot)$ as the cosine dissimilarity or the $\ell_2$ distance. However, when used as a learning objective we found the cosine dissimilarity to be more stable across experiments. We therefore used
$d(\cdot,\cdot) = 1 - \cos(\cdot,\cdot)$
throughout our experiments.

\paragraph{Visual transformer tokens.}
For the particular case of ViT-based encoders, we apply the alignment loss element-wise over tokens and then average. We do not average the tokens themselves. In addition, we consider only patch tokens and do not attempt to align CLS or register tokens.

\paragraph{Normalization.}
We apply normalization via an exponential moving average (EMA) of feature statistics maintained during training for the student and the teachers. 
%

\begin{figure}[t]
    \centering
    \begin{subfigure}[t]{0.48\linewidth}
        \centering
        \includegraphics[width=\linewidth]{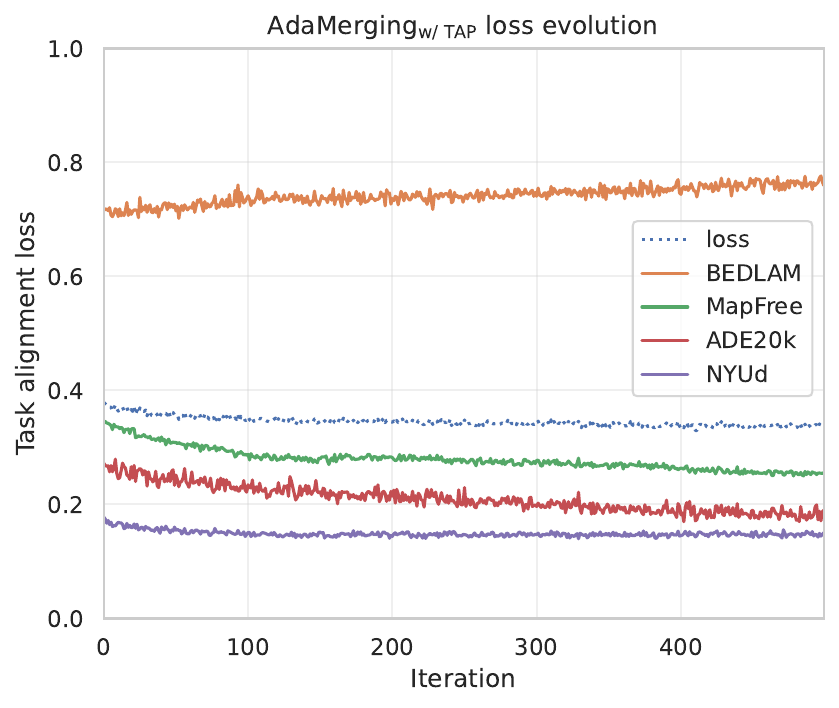}
        \label{fig:adamerge_loss}
    \end{subfigure}
    \hfill
    \begin{subfigure}[t]{0.48\linewidth}
        \centering
        \includegraphics[width=\linewidth]{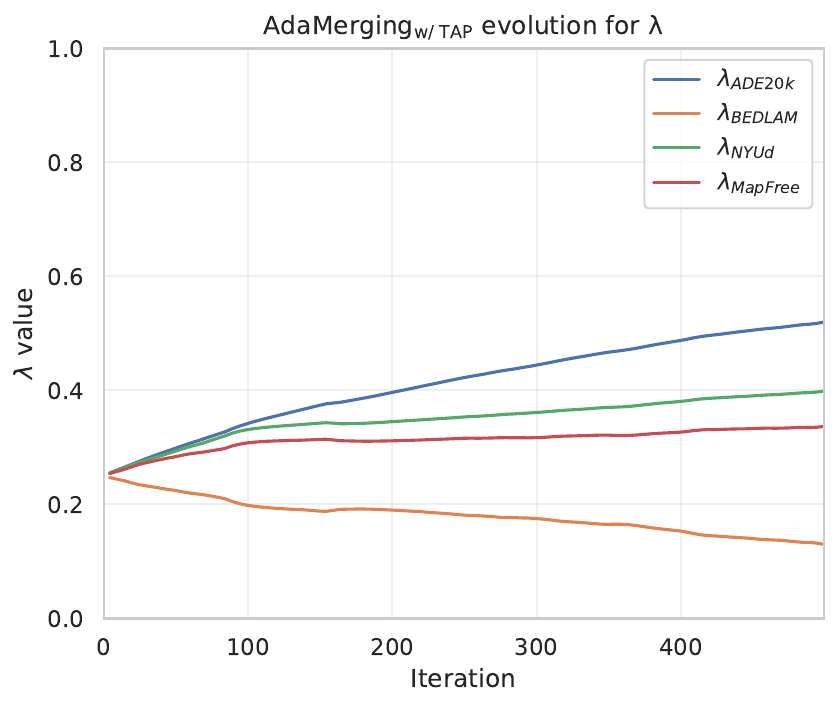}
        \label{fig:adamerge_lambdas}
    \end{subfigure}
    \caption{\textbf{Left:} Loss curves observed while training \tadamerge. We can observe similar behavior between dense tasks (\ade, \nyud), and a different behavior in \bedlam. \textbf{Right:} $\lambda$ values for different tasks observed while training \tadamerge. Again, we observe that similar tasks show a similar pattern, while more specific tasks (\bedlam) go a different path.}
    \label{fig:ada_figure}
\end{figure}

\paragraph{Task-specific structure.}
Feature-level alignment, unlike its use as a performance score, may have unintended effects. Task-specific encoders often allocate capacity to features that are only useful within their domain (\eg, MultiHMR).
For instance, models trained on data with strong spatial or semantic biases (such as datasets emphasizing specific human body parts for pose recovery) may produce ambiguous or noisy features in other regions. Our alignment loss does not differentiate between informative and less relevant dimensions, and may therefore encourage the student to replicate such task-specific artifacts.
This observation corroborates with our experiments, in which we found our \bedlam teachers to be most challenging to approximate (see~\cref{fig:ada_figure}).

\paragraph{Learning.}
For optimizing the merging coefficients, we follow the same settings as in~\cite{AdaMerging_ICLR_2024}. Specifically, we use the Adam optimizer \cite{kingma2014adam} with a learning rate of $1 \times 10^{-3}$, a batch size of 16, and run for 500 iterations. The merging coefficients $\lambda$ are initialized uniformly, \ie, $\lambda = 1 / T$ (\eg, $1/4 = 0.25$ in our case with four tasks).

\input{tables/tv_per_layer}

%% file: tables/appendix/lidar_datasets.tex
\begin{table*}[t]
    \centering
\caption{\textbf{Datasets used in our LiDAR model merging experiments.} For each dataset, we provide: reference, LiDAR sensor used for data capture, number of beams, vertical field of view (V.\,FoV), vertical resolution (V.\,res.), horizontal field of view (H.\,FoV), horizontal resolution (H.\,res.), number of classes used for standard benchmarking (which may be lower than the number of finer-grained actually annotated classes), number of frames for training and/or testing, and region of the world where the data was captured. Collection of information from \cite{michele2024ttyd}.}
\label{tab:lidar_datasets}
\vspace{-0.2cm}
\tabcolsep 0.8mm
\scalebox{0.62}{\begin{tabular}{lrlcrlllcrrl}
\toprule
         
Dataset   & Ref. & Lidar & \!\!\!\!\!Beams\!\!\! & \multicolumn{1}{c}{V.\,FoV} & V.\,res. & \multicolumn{1}{c}{H.\,FoV} & H.\,res. & \!\!\!\!\!Classes\!\! & Train & Test~ & Region of the world
\\
\midrule
nuScenes & \cite{caesar2020nuscenes,fong2021lidarseg_nuscenes} & Velodyne HDL-32E & 32 & -30.7° to +10.7° & 1.33° & 360° & 0.33° & 16 & {28,130} & 6,027 & Boston, Singapore
\\

SemanticKITTI & \cite{geiger2012cvpr, behley2019iccv} & Velodyne 
HDL-64E & 64 & -24.8° to +\hphantom{1}2.0° & 0.42° & 360° & 0.18° &  19 & 19,130 & 4,071 & Karlsruhe
\\
Panda64 & \cite{xiao2021pandaset} & Hesai Pandar64 & 64 & -25.0° to +15.0° & 0.17° &  360° &  0.20°/6° & 37 & 3,800 & {2,280} & San Francisco, \\
&&&&&&&&&&&El Camino Real
\\
PandaGT & \cite{xiao2021pandaset} & Hesai PandarGT & equiv. 150 & -10.0° to +10.0° & 0.07° & 60° & 0.10° & 37 & 3,800 & {2,280} & San Francisco, \\
&&&&&&&&&&&El Camino Real
\\

\bottomrule
\end{tabular}}

\end{table*}

%% file: floats/cost_vs_perf_lidar.tex
\begin{figure}[t]
    \centering
    \includegraphics[width=0.5\textwidth]{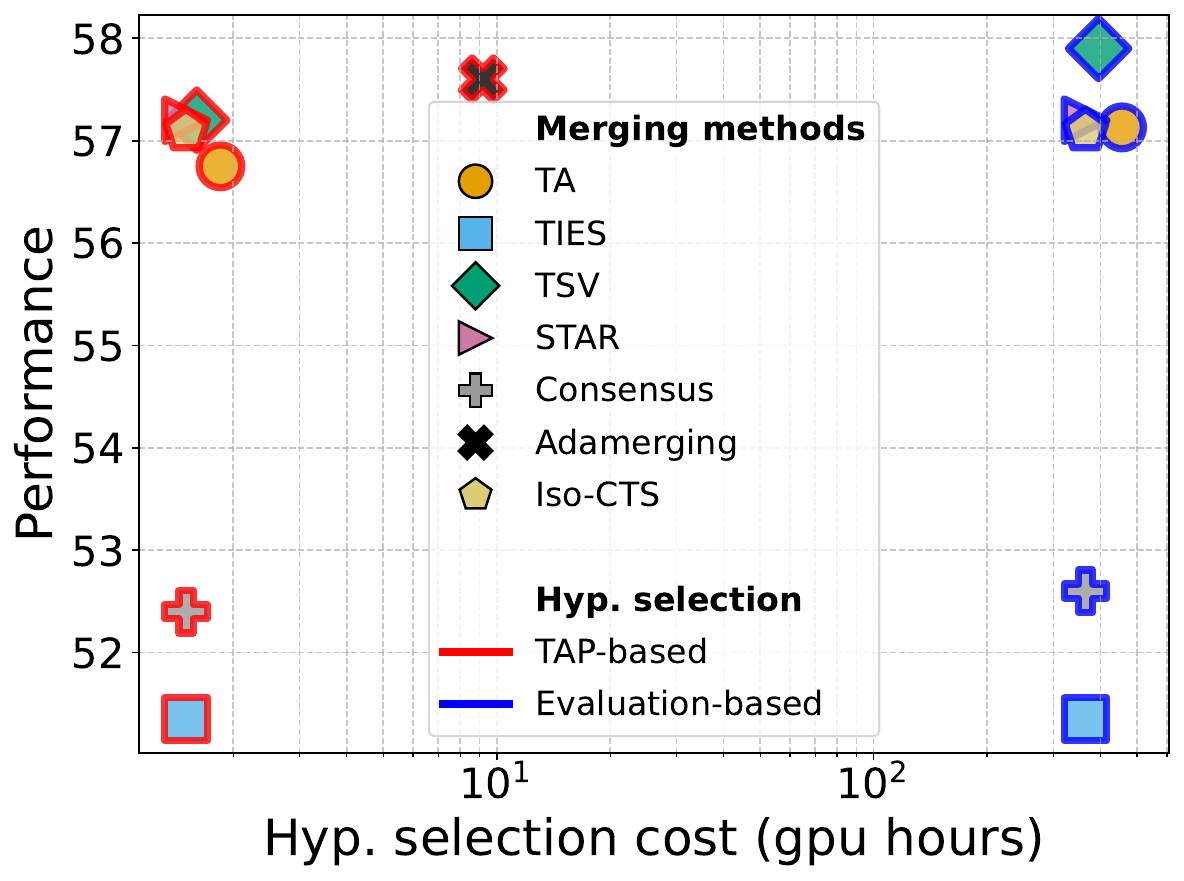}
    \caption{
      \textbf{Performance \vs Hyp. selection cost for LiDAR setting.} \talign reduces hyperparameter selection costs by 2 orders of magnitude when merging models for the LiDAR-based segmentation (see \cref{sec:protocol}). Note original Adamerging is not compatible with non-categorical tasks.}
    
    \label{fig:cost_vs_perf_lidar}
\end{figure}

%% file: tables/appendix/clip_hyp_search.tex
\begin{table*}[t]
\centering
\small
\setlength{\tabcolsep}{4pt} 
\caption{Hyperparameter search for CLIP setting (ViT-L, 20 tasks).}
\label{app:tab:clip_hyp_search}
\begin{tabular}{lll}
\toprule
Method & Hyperparameters Considered & Optimal Hparams \\
\midrule

TA
& $\lambda \in \{.05,.10,\ldots,1.00\}$
& $\lambda=.10$ \\

TIES
& $\lambda \in \{.77,.94,\ldots,1.8\}$, $K \in \{20,40,80\}$
& $\lambda=.94$, $K=80$ \\

TSV
& $\alpha \in \{.10,.20,\ldots,2.00\}$
& $\alpha=1.00$ \\

STAR
& $\eta \in \{.2,.4,.6,.8,.9,.95\}$, $\alpha \in \{.2,.4,\ldots,2.0\}$
& $\eta=.8$, $\alpha=1.8$ \\

Consensus
& $m_{\text{cons}}$ from paper, $\alpha \in \{.10,.20,\ldots,2.00\}$
& $\alpha=.4$ \\

Iso-CTS
& $\mathrm{csf}=.80$, $\alpha \in \{.15,.3,\ldots,3.0\}$
& $\alpha=1.2$ \\
\bottomrule
\end{tabular}
\end{table*}

%% file: tables/appendix/lidar_hyp_search.tex
\begin{table*}[t]
\centering
\small
\setlength{\tabcolsep}{4pt} 
\caption{Hyperparameter search during ablation for LiDAR setting.}
\label{app:tab:lidar_hyp_search}
\begin{tabular}{lll}
\toprule
Method & Hyperparameters Considered & Optimal Hparams \\
\midrule
TA
& $\lambda \in \{.025,.05,\ldots,.375\}$
& $\lambda=.15$ \\

TIES
& $\lambda \in \{.6,.8,1.0,1.2\}$, $K \in \{10,20,30\}$
& $\lambda=.6,\ K=30$ \\

TSV
& $\alpha \in \{.3,.4,\ldots,1.5\}$
& $\alpha=.5$ \\

STAR
& $\eta \in \{.6,.8,.9,.95\}$, $\alpha \in \{.8,1.0,1.2\}$
& $\eta=.8,\ \alpha=.8$ \\

Consensus
& $\lambda_t \in \{.2,.3,\ldots,.9\}$, $\alpha \in \{.1,.2,.3,.4,.6,.8,1.0,\ldots,2.0\}$
& $\lambda_t=.8,\ \alpha=.8$ \\

Iso-CTS
& $\mathrm{csf}=.8$, $\alpha \in \{.2,\ldots,2.0\}$
& $\mathrm{csf}=.8,\ \alpha=.6$ \\
\bottomrule
\end{tabular}
\end{table*}

%% file: tables/appendix/het_setting_hyp_search.tex
\begin{table*}[t]
\centering
\small
\setlength{\tabcolsep}{4pt} 
\caption{Hyperparameter search during ablation for Heterogeneous setting}
\label{app:tab:het_setting_hyp_search}
\begin{tabular}{lll}
\toprule
Method & Hyperparameters Considered & Optimal Hparams \\
\midrule
TA
& $\lambda \in \{.1,.2,\ldots,1.0\}$
& $\lambda=.4$ \\

TIES
& $\lambda \in \{.8,1.0,1.2\}$, $K \in \{5,10,20,30\}$
& $\lambda=.8,\ K=5$ \\

TSV
& $\alpha \in \{.6,.7,\ldots,1.5\}$
& $\alpha=.7$ \\

STAR
& $\eta \in \{.6,.8,.9,.95\}$, $\alpha \in \{.8,1.0,1.2\}$
& $\eta=.95,\ \alpha=1.0$ \\

Consensus
& $\lambda_t \in \{.2,.3,\ldots,.6\}$, $\alpha \in \{.1,.2,.3,.4,.6,.8,1.0\}$
& $\alpha=.6$ \\

Iso-CTS
& $\mathrm{csf}=.8$, $\alpha \in \{.2,.4,\ldots,2.0\}$
& $\mathrm{csf}=.8,\ \alpha=.2$ \\
\bottomrule
\end{tabular}
\end{table*}

%% file: tables/appendix/selected_hyps_long_eval_het_setting.tex
\begin{table*}[t]
\centering
\small
\setlength{\tabcolsep}{4pt}
\caption{Selected hyperparameters for main results in \cref{tab:dune} and \cref{tab:dinov2}}
\label{app:tab:selected_long_eval}
\begin{tabular}{lll}
\toprule
Method & Selected Hyps (\cref{tab:dune}) & Selected Hyps (\cref{tab:dinov2}) \\
\midrule
TA 
& $\lambda=0.25$ 
& $\lambda=.1$ \\

TIES 
& $\lambda=.2,\ K=5$ 
& $\lambda=.2,\ K=5$ \\

TSV 
& $\lambda=.8$ 
& $\lambda=.3$ \\

STAR 
& $\eta=.8,\ \lambda=.8$ 
& $\eta=.99$,\ $\lambda=.8$  \\

Consensus 
& $\lambda=.8$ 
& $\lambda=.3$ \\

Iso-CTS 
& $\lambda=.18,\ \mathrm{csf}=.0001$ 
& $\lambda=.14,\ \mathrm{csf}=.0001$ \\

\bottomrule
\end{tabular}
\end{table*}

%% file: tex/extended_rel_work_v1.tex
In this section, we discuss in more detail how each merging method can be expressed as an instantiation of \cref{eqn:th} introduced in \cref{sec:related}, that is:
\begin{equation}
\theta_{\text{merged}}(\lambda, \mu) = \theta_0 + \sum_{t=1}^T \lambda_t \odot \phi(\tau_t; \mu),
\tag{\ref{eqn:th}}
\end{equation}
where $\lambda_t \in \mathbb{R}^d$ is a vector of task-specific weights controlling \emph{how much} of each task vector is added to the base model (if $\lambda_t = \mathbf{0}$ for all $t$, the merged model reduces to $\theta_0$), $\phi: \mathbb{R}^d \rightarrow \mathbb{R}^d$ is a transformation of the task vectors parametrised by hyperparameters $\mu$, and $\odot$ denotes the Hadamard product.

The two hyperparameter sets play distinct roles: $\lambda$ governs the \emph{contribution} of each task vector to the merged model, while $\mu$ governs the \emph{structure} of the task vectors themselves before merging.
When $\phi$ does not involve any free hyperparameters (\eg it is the identity or a fixed deterministic function of the task vectors), $\mu$ is the empty set. 

Note that for simplicity, we keep the notation $\phi(\tau_t;\mu)$ but for some methods whose transformations depend on the whole set of task vectors, we allow $\mu$ to include masks, signs, or bases computed from $\{\tau_j\}_{j=1}^T$. A more correct formulation could be to express $\phi$ as $\phi(\tau_1,...,\tau_T; \mu, t)$. That is, $\phi$ depends on all task vectors and potentially yields a different value for each task, in the simplest cases only depending on $\tau_t$.

We now instantiate this framework for each method surveyed in \cref{sec:related}, and summarize the result in \cref{tab:merging_unified}.

\paragraph{Model averaging.}
The task-specific weight is a uniform scalar, $\lambda_t = \tfrac{1}{T}\mathbf{1}$, identical across all tasks and all parameters.
The transformation is the identity, $\phi(\tau_t) = \tau_t$, so that $\theta_{\text{merged}} = \theta_0 + \tfrac{1}{T}\sum_t \tau_t = \tfrac{1}{T}\sum_t \theta_t$.
Since $\phi$ is the identity, $\mu = \emptyset$.

\paragraph{Task Arithmetic~\cite{ilharco2023editing}.}
A single global scalar $\lambda_t = \lambda\,\mathbf{1}$ (shared across tasks and parameters) controls how far the merged model moves from the base.
Again $\phi$ is the identity and $\mu = \emptyset$.
The sole hyperparameter is therefore $\lambda$.

\paragraph{TIES~\cite{prateek2023ties}.}
The contribution of each task is again governed by a single shared scalar, $\lambda_t = \lambda\,\mathbf{1}$.
However, the transformation $\phi$ is non-trivial and carries all the method-specific structure via $\mu = K$, where $K$ is a top-$K$ trimming fraction.
Given $K$ and the full collection $\{\tau_t\}$, $\phi$ proceeds in two steps: (i)~\emph{trim}---set to zero all entries of $\tau_t$ whose magnitude falls below the top-$K$ threshold, producing a sparse vector $\hat{\tau}_t$; and (ii)~\emph{elect}---for each parameter dimension, retain only the entries of $\hat{\tau}_t$ whose sign agrees with the aggregate sign $\operatorname{sgn}\!\left(\sum_t \hat{\tau}_t\right)$, zeroing out the rest.

\paragraph{Breadcrumbs~\cite{davari2023model}.}
Similarly to TIES, $\lambda_t = \lambda\,\mathbf{1}$ and $\phi$ applies sparse masking, with $\mu = (\alpha, \beta)$.
However, the mask is computed at the \emph{layer} level: for each layer $\ell$, $\phi$ removes both the lowest-magnitude weights (below a fraction $\alpha$ of the layer norm) and the highest-magnitude weights (above a fraction $1-\beta$), retaining only the intermediate range and thus eliminating outliers that could destabilise merging.

\paragraph{Consensus~\cite{wang2024localizing}.}
Again $\lambda_t = \lambda\,\mathbf{1}$ and $\mu = k$.
The transformation $\phi$ first builds a binary mask $m_t$ for each task by retaining the entries fulfilling $|\tau_t| \geq k |\sum_t \tau_t - \tau_t|$. That is, intuitively, only keep those weight updates that bring the sum of task vectors ($\sum_t \tau_t$) closer to the fine-tuned model for task $t$. A tolerance hyperparameter $k$ with "misaligned" weights is ablated for each task. 
It then forms a \emph{consensus mask}, $m_{\text{consensus}} = \mathbf{1}\!\left[\sum_t m_t \geq k\right]$, that retains only parameters present in at least two individual masks, and applies it: $\phi(\tau_t;\mu) = m_{\text{consensus}} \odot \tau_t$.

\paragraph{Lines~\cite{wang2025lines}.}
Here $\phi$ is the identity ($\mu = \emptyset$), so all structure is encoded in $\lambda_t$.
The per-parameter weight is constant within each transformer layer $\ell$ but varies \emph{across} layers: $(\lambda_t)_\ell = \lambda_0 + \ell \cdot \Delta\lambda$, where $\lambda_0$ and $\Delta\lambda$ are shared scalars.
This linear schedule assigns smaller coefficients to early (more general) layers and larger ones to later (more task-specific) layers.

\paragraph{STAR~\cite{lee2025star}.}
For matrix-shaped weight tensors, $\phi$ applies a truncated SVD with $\mu = \eta$, where $\mu$ is a target fraction of spectral energy to retain.
Let $\tau_t^{(\ell)} = U S V^\top$ be the SVD of the weight-matrix-shaped slice of $\tau_t$ at layer $\ell$.
The transformation retains only the top-$r$ singular values chosen so that $\|S_r\|_{tr} / \|S\|_{tr} \geq \mu$, where $\|\cdot\|_{tr}$ denotes the trace norm. Then, it rescales the remaining singular values to match the original trace norm: $\phi(\tau_t^{(\ell)};\mu) = \sigma \, U_r S_r V_r^\top$, where $\sigma = \|S\|_{tr} / \|S_r\|_{tr}$.
For non-matrix tensors, $\phi$ reduces to the identity.
The contribution of each task is governed by a global scalar $\lambda_t = \lambda\,\mathbf{1}$.

\paragraph{TSV~\cite{gargiulo2025task}.}
TSV enforces inter-task orthogonality, with $\mu = 1/T$ controlling the truncation rank.
For each matrix-shaped layer $\ell$, $\phi$ first performs an individual truncated SVD on each task vector (as in STAR), yielding low-rank factors $(U_t^{(\ell)}, S_t^{(\ell)}, V_t^{(\ell)})$.
It then concatenates the left singular-vector matrices across tasks, $\mathcal{U}^{(\ell)} = [U_1^{(\ell)} \mid \cdots \mid U_T^{(\ell)}]$, and applies a second SVD to $\mathcal{U}^{(\ell)}$ to obtain a shared orthonormal basis, similarly for the right singular vectors. Then, the orthonormal left and right singular vectors across tasks are combined with the block diagonal matrix of joint $S_t^{(\ell)}$.
As with STAR, $\lambda_t = \lambda\,\mathbf{1}$ and $\phi$ defaults to the identity for non-matrix weights.

\paragraph{Iso Merging~\cite{marczak2025notaskleftbehind}.} Isotropic merging relies on a single scaling factor $\lambda_t = \lambda\,\mathbf{1}$ and similar to STAR and TSV performs SVD decomposition on the weight matrices. However, it decomposes not only the task vectors $\tau_t$ but also the "merged" task vector $\tau_{\text{TA}} = \sum_t \tau_t$. It introduces two flavours of merging, in the simplest one, Iso-C, $\phi$ performs SVD on $\tau_{\text{TA}}^{(\ell)} = (U^{(\ell)}, S^{(\ell)}, V^{(\ell)})$ and sets all singular values of $S^{(\ell)}$ to $\bar{\sigma} = 1/k \sum \sigma_k$. On the other hand Iso-CTS adds an additional hyperparameter $\mu = \text{cfs}$ which is denoted as the Common Space Fraction. Based on this, $\phi$ concatenates the truncated SVD's of \textit{task specific} spaces $\tau_t^{(\ell)}$ and  the \textit{common} space $\tau_{\text{TA}}^{(\ell)}$ in a similar fashion as done in TSV and the hyperparameter csf is used to control the ratio of the truncated ranks for the common and task specific fractions. Moreover, similar to Iso-C, all singular values are set to the average value.

\paragraph{AdaMerging~\cite{AdaMerging_ICLR_2024}.}
Unlike the methods above, AdaMerging \emph{learns} $\lambda$ rather than fixing it, so $\lambda$ itself is the output of an optimisation rather than a hyperparameter in the usual sense.
The transformation $\phi$ is the identity ($\mu = \emptyset$), but $\lambda_t \in \mathbb{R}^d$ is optimised by minimising the entropy of the merged model's predictions on unlabelled test data.
In the layer-wise variant, $\lambda_t$ is constant within each layer; in the task-wise variant, it is a single scalar per task.

\paragraph{AdaMMS~\cite{du2025adamms}.}
AdaMMS rather than a merging method in itself, is actually a hyperparameter selection method. It selects $\lambda$ by measuring the \emph{generation consistency} of models obtained at adjacent hyperparameter values $\lambda$ and $\lambda + \epsilon$, however, due to this design, it only works with methods with a single scalar free parameter, that is, $\lambda_t = \lambda\,\mathbf{1}$ and $\mu = \emptyset$.

\begin{table}[t]
\centering
\caption{%
  Instantiation of the unified merging framework of \cref{eqn:th}
  for each method discussed in \cref{sec:related}.
}
\label{tab:merging_unified}
\resizebox{\linewidth}{!}{%
\begin{tabular}{llll}
\toprule
\textbf{Method} & \textbf{Task weight} $\lambda_t$ & \textbf{Transformation} $\phi(\tau_t;\mu)$ & \textbf{Transf.\ hyperparameters} $\mu$ \\
\midrule
Model Averaging
  & $\frac{1}{T}\mathbf{1}$
  & Identity
  & $\emptyset$ \\[4pt]
Task Arithmetic~\cite{ilharco2023editing}
  & $\lambda\,\mathbf{1}$
  & Identity
  & $\emptyset$ \\[4pt]
TIES~\cite{prateek2023ties}
  & $\lambda\,\mathbf{1}$
  & Top-$K$ trim $+$ sign-based election
  & $K$-percentage of kept weights  \\[4pt]
Breadcrumbs~\cite{davari2023model}
  & $\lambda\,\mathbf{1}$
  & Layer-wise mask (remove low \& high magnitude)
  & $\alpha,\,\beta$ top and bottom thresholds\\[4pt]
Consensus~\cite{wang2024localizing}
  & $\lambda\,\mathbf{1}$
  & Per-task mask + consensus across tasks
  & $k$ tolerance with misaligned weights \\[4pt]
Lines~\cite{wang2025lines}
  & $(\lambda_0 + \ell\cdot\Delta\lambda)\,\mathbf{1}_\ell$
  & Identity
  & $\emptyset$ \\[4pt]
STAR~\cite{lee2025star}
  & $\lambda\,\mathbf{1}$
  & weight matrix SVD $+$ $\sigma$ rescaling
  & $\eta$ percentage of trace norm to keep \\[4pt]
TSV~\cite{gargiulo2025task}
  & $\lambda\,\mathbf{1}$
  & weight matrix SVD + joint orthogonal basis
  & $\emptyset$ \\[4pt]
Iso-CTS~\cite{marczak2025notaskleftbehind}
  & $\lambda\,\mathbf{1}$
  & SVD + joint orthogonal basis + $\sigma$ rescaling
  & csf relative truncation threshold \\[4pt]
AdaMerging~\cite{AdaMerging_ICLR_2024}
  & learned $\lambda_t$ (task/layer-wise)
  & Identity
  & $\emptyset$ \\[4pt]
AdaMMS~\cite{du2025adamms}
  & $\lambda\,\mathbf{1}$ (auto-selected)
  & Identity
  & $\emptyset$ \\
\bottomrule
\end{tabular}%
}
\end{table}


%% file: tables/appendix/clip_full.tex
\begin{table}[t]
\caption{ \textbf{Hyperparameter vs \talign selection on CLIP benchmark}. We report the test performance of the merged models with exhaustive hyperparameter selection on the validation set \vs \talign selected hyperparameters. For each merging method, best results are in bold. Despite not having access to labels and using only a fraction of the validation images, the drop in performance with \talign selection is minimal. Moreover, for \adamerge, it seems that optimizing \talign is more robust than entropy as we increase the number of classes.
}
\centering
\label{tab:clip_full}
\begingroup
\scriptsize
\setlength{\tabcolsep}{3pt}

\begin{tabular}{l ccc @{\hspace{10pt}} ccc @{\hspace{10pt}} ccc @{\hspace{10pt}} c}
\toprule
& \multicolumn{3}{c}{\texttt{ViT-B-32}} 
& \multicolumn{3}{c}{\texttt{ViT-B-16}} 
& \multicolumn{3}{c}{\texttt{ViT-L-14}} 
& \multirow{2}{*}{Average} \\

\cmidrule(lr){2-4}
\cmidrule(lr){5-7}
\cmidrule(lr){8-10}

& T=8 & T=14 & T=20 
& T=8 & T=14 & T=20 
& T=8 & T=14 & T=20 
&  \\

\midrule

Zero-shot    & 48.3 & 57.2 & 56.1 & 55.3 & 61.3 & 59.7 & 64.7 & 68.2 & 65.2 & 59.6 \\
Fine-tuned   & 92.8 & 90.9 & 91.3 & 94.6 & 92.8 & 93.2 & 95.8 & 94.3 & 94.7 & 93.4 \\
\midrule 
\rowcolor{gray!10}
TA$_\text{w/\HS}$ & {\scriptsize {71.14}} & {\scriptsize {65.19}} & {\scriptsize {62.61}} 
& {\scriptsize {76.18}} & {\scriptsize {70.44}} & {\scriptsize {68.19}} 
& {\scriptsize {84.90}} & {\scriptsize 79.41} & {\scriptsize {76.01}} 
& {\scriptsize {72.68}} \\

\rowcolor{gray!10}
TA$_{\text{w/\talign}}$ & {\scriptsize 71.13} & {\scriptsize 65.02} & {\scriptsize 62.07} 
& {\scriptsize 75.25} & {\scriptsize {70.44}} & {\scriptsize {68.19}} 
& {\scriptsize {84.90}} & {\scriptsize {79.69}} & {\scriptsize {76.01}} 
& {\scriptsize 72.52} \\

\addlinespace[0.6ex]

TIES$_\text{w/\HS}$ & {\scriptsize {74.76}} & {\scriptsize {67.80}} & {\scriptsize {64.75}} 
& {\scriptsize {79.82}} & {\scriptsize {73.08}} & {\scriptsize {70.02}} 
& {\scriptsize 86.87} & {\scriptsize {80.17}} & {\scriptsize {77.09}} 
& {\scriptsize {74.93}} \\

TIES$_{\text{w/\talign}}$ & {\scriptsize 74.50} & {\scriptsize {67.80}} & {\scriptsize {64.75}} 
& {\scriptsize {79.82}} & {\scriptsize {73.08}} & {\scriptsize {70.02}} 
& {\scriptsize {87.02}} & {\scriptsize {80.17}} & {\scriptsize {77.09}} 
& {\scriptsize 74.92} \\

\addlinespace[0.6ex]

\rowcolor{gray!10}
TSV$_\text{w/\HS}$ & {\scriptsize {85.77}} & {\scriptsize {80.03}} & {\scriptsize {76.93}} 
& {\scriptsize {89.01}} & {\scriptsize 84.28} & {\scriptsize {80.65}} 
& {\scriptsize {93.00}} & {\scriptsize {89.20}} & {\scriptsize 87.52} 
& {\scriptsize {85.15}} \\

\rowcolor{gray!10}
TSV$_{\text{w/\talign}}$ & {\scriptsize 85.50} & {\scriptsize {80.03}} & {\scriptsize 76.88} 
& {\scriptsize {89.01}} & {\scriptsize {84.52}} & {\scriptsize 80.45} 
& {\scriptsize 92.99} & {\scriptsize 89.12} & {\scriptsize {87.55}} 
& {\scriptsize 85.12} \\

\addlinespace[0.6ex]

STAR$_\text{w/\HS}$ & {\scriptsize {71.39}} & {\scriptsize {65.42}} & {\scriptsize 62.94} 
& {\scriptsize {76.32}} & {\scriptsize {70.70}} & {\scriptsize {68.49}} 
& {\scriptsize {84.86}} & {\scriptsize {79.73}} & {\scriptsize {76.08}} 
& {\scriptsize {72.88}} \\

STAR$_{\text{w/\talign}}$ & {\scriptsize 70.90} & {\scriptsize 65.34} & {\scriptsize {62.98}} 
& {\scriptsize 76.17} & {\scriptsize 70.63} & {\scriptsize 68.44} 
& {\scriptsize {84.86}} & {\scriptsize {79.73}} & {\scriptsize {76.08}} 
& {\scriptsize 72.79} \\

\addlinespace[0.6ex]

\rowcolor{gray!10}
Consensus$_\text{w/\HS}$ & {\scriptsize {74.91}} & {\scriptsize {70.33}} & {\scriptsize {66.94}} 
& {\scriptsize {79.30}} & {\scriptsize {74.33}} & {\scriptsize {71.76}} 
& {\scriptsize {86.33}} & {\scriptsize {82.31}} & {\scriptsize {79.63}} 
& {\scriptsize {76.20}} \\

\rowcolor{gray!10}
Consensus$_{\text{w/\talign}}$ & {\scriptsize 74.69} & {\scriptsize 70.32} & {\scriptsize {66.94}} 
& {\scriptsize {79.30}} & {\scriptsize {74.33}} & {\scriptsize {71.76}} 
& {\scriptsize {86.33}} & {\scriptsize {82.31}} & {\scriptsize {79.63}} 
& {\scriptsize 76.18} \\

\addlinespace[0.6ex]

Iso-CTS$_\text{w/\HS}$ & 86.48 & 81.56 & 78.17 & 91.16 & 86.29 & 82.81 & 94.81 & 91.02 & 90.19 & 86.94 \\
Iso-CTS$_{\text{w/\talign}}$ & {\scriptsize 85.77} & 81.56 & {\scriptsize 78.08} & {\scriptsize 91.03} & 86.29 & {\scriptsize 82.57} & {\scriptsize 94.72} & 91.02 & 90.19 & {\scriptsize 86.80} \\

\addlinespace[0.6ex]

\rowcolor{gray!10}
AdaMerging$_{\text{w/Entropy}}$ & {\scriptsize {84.15}} & {\scriptsize 74.97} & {\scriptsize 71.49} 
& {\scriptsize {84.39}} & {\scriptsize 77.43} & {\scriptsize 74.96} 
& {\scriptsize 91.42} & {\scriptsize 87.02} & {\scriptsize 84.34} 
& {\scriptsize 81.13} \\

\rowcolor{gray!10}
AdaMerging$_{\text{w/\talign}}$ & {\scriptsize 80.72} & {\scriptsize {75.36}} & {\scriptsize {72.35}} 
& {\scriptsize 83.12} & {\scriptsize {78.88}} & {\scriptsize {76.36}} 
& {\scriptsize {91.73}} & {\scriptsize {87.68}} & {\scriptsize {87.05}} 
& {\scriptsize {81.47}} \\

\bottomrule
\end{tabular}
\endgroup
\end{table}

%% file: tables/appendix/lidar_table_full.tex
\begin{table}[t]
\caption{
\textbf{Merging LiDAR models.} Results after merging different 3D segmentation models with \talign-based hyperparameter selection. We successfully apply several methods, significantly improving over the pretrained baseline. We report the mIoU, the average mIoU and the normalized performance w.r.t. the fine-tuned models.
%
}
\label{tab:app:lidar_full}
\centering
\begin{tabular*}{\textwidth}{@{\extracolsep{\fill}}lrrrrrr}
\toprule

\makecell{ \\ } & \makecell{nuScenes \\ mIoU (\(\uparrow\))} & \makecell{Sem.KITTI \\ mIoU (\(\uparrow\))} & \makecell{Panda64 \\ mIoU (\(\uparrow\))} & \makecell{PandaGT \\ mIoU (\(\uparrow\))} & \makecell{Avg. \\ mIoU(\(\uparrow\))} & \makecell{Normalized \\ Performance} \\
\midrule
Pretrained & 68.4 & 55.6 & 37.0 & 35.4 & 49.1 & 0.841\\
\midrule
TA$_{\text{w/\HS}}$ & 74.5 &	63.4 &	46.6 &44.0 &	57.1	&	0.978 \\
TA$_{\text{w/\talign}}$ & 73.7 & 63.5 & 47.0 & 42.9 & 56.8 & 0.972 \\
\midrule
STAR$_\text{w/\HS}$ &74.1 &	63.7 &	47.5 &	43.3 &	57.2	& 0.979  \\
STAR$_{\text{w/\talign}}$ & 74.1 & {63.7} & {47.5} & 43.3 & 57.2 &  {0.979} \\
\midrule
TIES$_{\text{w/\HS}}$ & 69.0 & 	56.0 &	41.9 &	38.5 &	51.4 &	0.879\\
TIES$_{\text{w/\talign}}$ & 69.0 & 56.0 &41.9 &38.5 & 51.4 & 0.879\\
\midrule
Consensus$_{\text{w/\HS}}$ & 70.8 & 	59.8 &	41.1 &	38.8 &	52.6 &	0.901\\
Consensus$_{\text{w/\talign}}$ & 70.1 & 59.5 & 41.3 & 38.8  & 52.4 &  0.898\\
\midrule
TSV$_{\text{w/\HS}}$ & 75.6	& 64.0	& 47.5 & 44.7 & 58.0 & 0.992 \\
TSV$_{\text{w/\talign}}$ & {74.7} & 63.5 & 47.2 & 43.4 & 57.2 & 0.979 \\
\midrule
Iso-CTS$_{\text{w/\HS}}$ & 74.8 & 62.4 & 47.0 & 44.6	& 57.2 & 0.979\\
Iso-CTS$_{\text{w/\talign}}$ & 74.8 & 62.4 & 47.0 & 44.6	& 57.2 & 0.979\\
\midrule
AdaMerging$_{\text{w/\talign}}$ & 73.1 & {64.4} & 46.6 & {46.1} & {57.6}& {0.985} \\
\midrule
Finetuning & 78.4 &	65.8 &	48.3 &	41.1 & 58.4 & 1.000 \\
\bottomrule
\end{tabular*}

\end{table}

%% file: floats/task_combinations_lidar.tex
\begin{figure}[t]
  \centering
  
  \begin{tikzpicture}
    \begin{axis}[
      scale only axis,
      width=0.37\textwidth,
      height=4cm,
      ybar,
      ymin=90, ymax=104,
      ylabel={\small Normalized Performance (\%)},
      xtick={0,1,2,3},
      xticklabels={
        {\makecell[l]{PD64+\\PDGT}},
        {\makecell[l]{PD64\\NS}},
        {\makecell[l]{PDGT\\NS}},
        {\makecell[l]{PD64\\PDGT\\NS}}
      },
      xticklabel style={rotate=0, xshift=0pt, font=\tiny},
      xtick style={line width=2pt, color=black},
      xtick pos=bottom,
      bar width=0.16,            
      enlarge x limits=0.2,     
      grid=major,
      major grid style={dashed,gray!50},
      area legend,
      legend style={
        at={(0.98,0.98)},        
        anchor=north east,       
        inner sep=2pt,           
        font=\tiny,
        fill=white, fill opacity=1, draw opacity=1,text opacity=1,
        legend cell align=left   
      }
    ]

    \addplot+[draw=RoyalBlue, fill=RoyalBlue]   
    coordinates { (0,102.1) (1,97.0) (2,98.2) (3,97.7) };
    \addplot+[draw=Black, fill=Black]    
    coordinates { (0, 102.0) (1, 96.5) (2,98.1) (3,97.0) };
    \addplot+[draw=Maroon, fill=Maroon]  
    coordinates { (0,101.1) (1,97.6) (2,99.2) (3,99.5) };

    \legend{STAR, TA,
    TSV}
    \end{axis}
  \end{tikzpicture}
  \caption{
  \textbf{Performance \vs task subsets} in the LiDAR setting using nuScenes~(NS), Panda64~(PD64) and PandaGT~(PDGT). When task vector norms are small and balanced (left-most), merging methods perform significantly better than when some task vectors are much larger than others.}
  \label{fig:app:task_combinations_lidar}
 
\end{figure}

%% file: floats/dinov2_ablation.tex
\begin{figure}[t]
    \centering
    \begin{minipage}{0.47\textwidth}
        \centering
        \includegraphics[width=\linewidth]{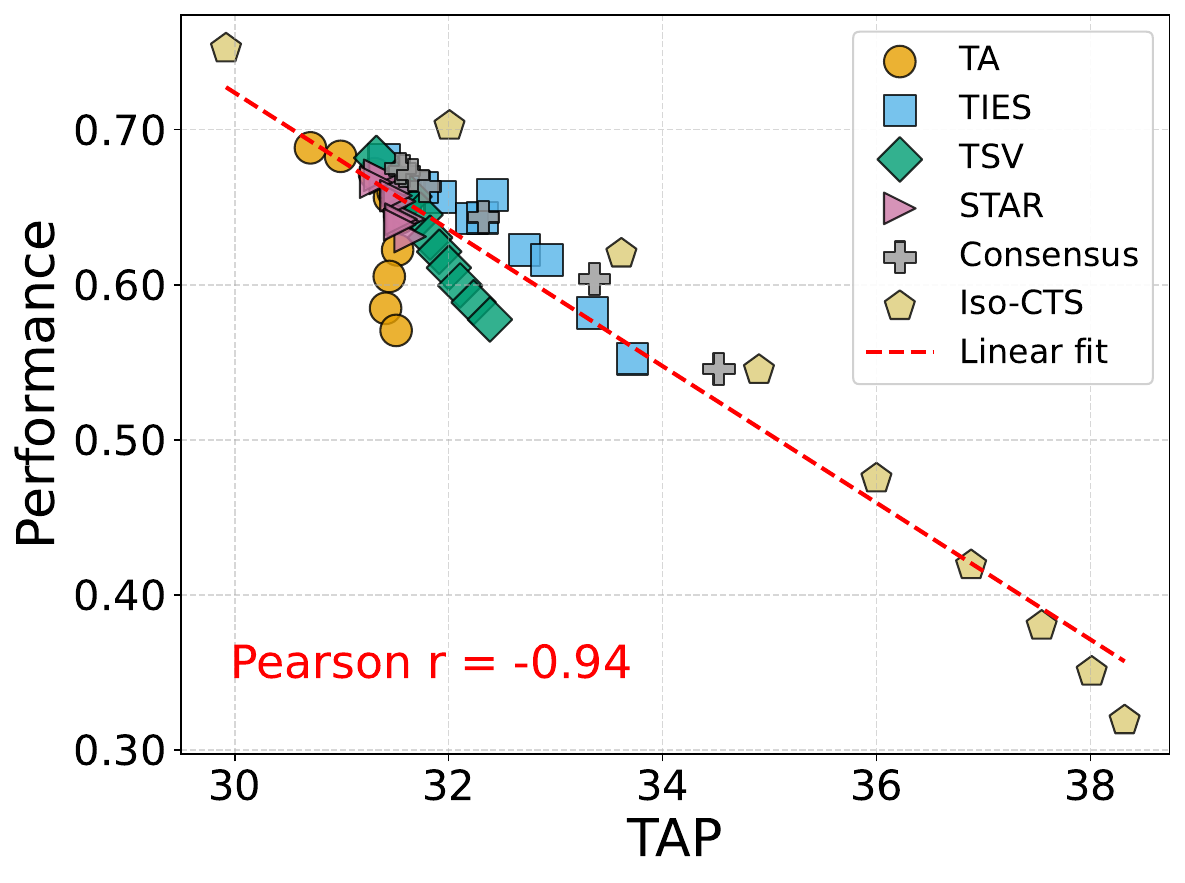}
    \end{minipage}
    \begin{minipage}{0.47\textwidth}
        \centering
        \includegraphics[width=\linewidth]{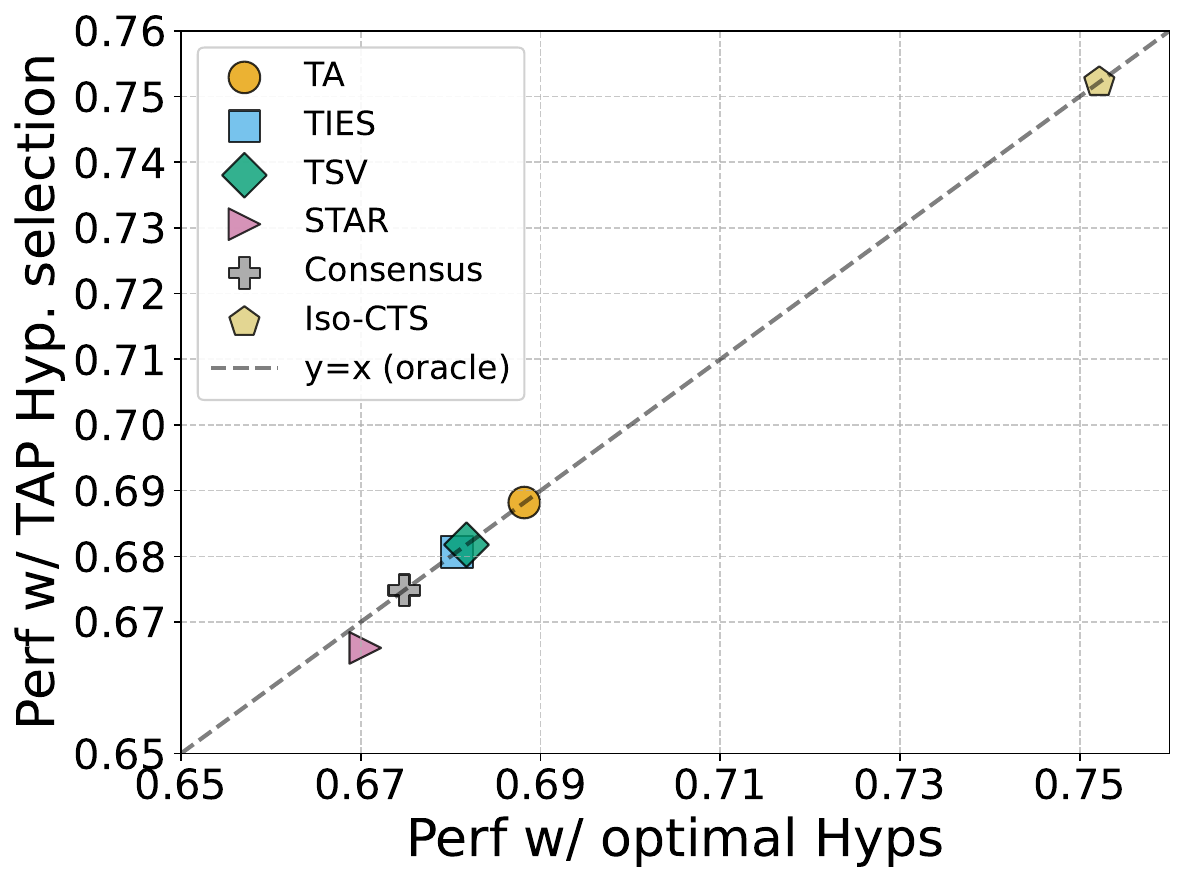}
    \end{minipage}
    \caption{
    \textbf{Left:} Correlation between Task Alignment Proxy (TAP) and Performance
on the validation sets when merging models in the heterogeneous vision tasks with DINOv2 as the base model. \textbf{Right:} Performance with TAP-selected hyper-
parameters vs. optimal hyperparameters (with costly downstream evaluation) for the
same setting. In line with \cref{fig:talign_scatters}, we observe a strong correlation between TAP and performance, which
leads to almost optimal hyperparameter selection without any downstream task evaluation. 
}
    \label{fig:app:dinov2_ablation}
\end{figure}

%% file: fig/param_change_per_layer.tex
\begin{tikzpicture}
\begin{axis}[
    width=0.8\linewidth, height=0.5\linewidth,
    ymin=0.4, ymax=80,              
    ymode=log,
    ytick={0.2,0.6,1,10,20},
    yticklabels={0.2,0.6,1,10,20},
    minor y tick num=9,
    xtick={1,...,12},
    xlabel={ViT-B Encoder Layer},
    ylabel={Parameter change (log scale)},
    grid=both,
    ymajorgrids=true, yminorgrids=true,
    legend cell align={left},
    legend style={
        font=\footnotesize,
        draw=black, fill=white,
        at={(1.4,0.5)}, anchor=east,
        legend columns=1
    }
]

\addplot+[
    black, mark=*, very thick, mark size=2.2pt,
    error bars/.cd, y dir=both, y explicit
] coordinates {
    (1, 0.665) +- (0, 0.073)
    (2, 0.640) +- (0, 0.073)
    (3, 0.620) +- (0, 0.081)
    (4, 0.630) +- (0, 0.071)
    (5, 0.645) +- (0, 0.080)
    (6, 0.655) +- (0, 0.097)
    (7, 0.655) +- (0, 0.116)
    (8, 0.645) +- (0, 0.116)
    (9, 0.675) +- (0, 0.109)
    (10, 0.730) +- (0, 0.135)
    (11, 0.760) +- (0, 0.143)
    (12, 0.730) +- (0, 0.187)
};
\addlegendentry{CLIP-mean/std}

\addplot+[
    MidnightBlue, mark=*, very thick, mark size=2.2pt,
    mark options={solid}
] coordinates {
 (1,1.9) (2,1.8) (3,1.6) (4,2.5) (5,2.8) (6,3.1)
    (7,3.2) (8,3.3) (9,3.4) (10,3.3) (11,3.2) (12,3.3)
};
\addlegendentry{DUNE-\nyud}

\addplot+[
    Goldenrod, mark=*, very thick, mark size=2.2pt,
    mark options={solid}
] coordinates {
    (1,3.5) (2,3.4) (3,3.6) (4,4.6) (5,4.5) (6,4.7)
    (7,5.0) (8,5.4) (9,5.8) (10,6.3) (11,6.8) (12,6.6)
};
\addlegendentry{DUNE-\ade}

\addplot+[
    BrickRed, mark=*, very thick, mark size=2.2pt,
    mark options={solid}
] coordinates {
    (1,22.7) (2,26.7) (3,23.1) (4,23.0) (5,22.6) (6,22.8)
    (7,23.4) (8,23.9) (9,24.5) (10,25.1) (11,25.4) (12,25.7)
};
\addlegendentry{DUNE-\mapfree}

\addplot+[
    OliveGreen, mark=*, very thick, mark size=2.2pt,
    mark options={solid}
] coordinates {
    (1,47.1) (2,50.4) (3,52.2) (4,51.9) (5,54.7) (6,57.2)
    (7,58.7) (8,60.5) (9,61.4) (10,63.2) (11,66.5) (12,69.5)
};
\addlegendentry{DUNE-BEDLAM}

\end{axis}
\end{tikzpicture}

%% file: tables/tv_per_layer.tex
\begin{table}[t]
\centering
\caption{
Per-layer task vector norms for the CLIP and DUNE benchmarks with the ViT-Base architecture.
{\em Pos.}: positional embeddings,
{\em CLS}: the CLS token,
{\em Conv.}: the first convolutional layer producing patch embeddings,
{\em Layer-i}: the i-th Transformer block,
{\em Norm}: the final LayerNorm.
We omit the comparison of CLIP parameters (the pre-LayerNorm and post-projection layers) which do not have correspondences in the DUNE-based ViT models.
}
\label{tab:tv_per_layer}
\adjustbox{max width=\linewidth}{
\begin{tabular}{l*{17}{r}}
Model & Pos. & CLS & Conv. & Layer-1 & Layer-2 & Layer-3 & Layer-4 & Layer-5 & Layer-6 & Layer-7 & Layer-8 & Layer-9 & Layer-10 & Layer-11 & Layer-12 & Norm & Avg. \\ 
\toprule
\multicolumn{15}{l}{\em CLIP benchmark} \\
Cars  & 0.1 & 0.0 & 0.1 & 0.7 & 0.8 & 0.7 & 0.7 & 0.7 & 0.7 & 0.6 & 0.6 & 0.7 & 0.8 & 0.9 & 1.0 & 0.0 & 0.6 \\
CIFAR10 & 0.1 & 0.0 & 0.1 & 0.7 & 0.6 & 0.6 & 0.6 & 0.6 & 0.7 & 0.7 & 0.6 & 0.7 & 0.7 & 0.7 & 0.6 & 0.0 & 0.5 \\
CIFAR100 & 0.1 & 0.0 & 0.1 & 0.7 & 0.6 & 0.6 & 0.6 & 0.7 & 0.7 & 0.7 & 0.8 & 0.8 & 0.9 & 0.9 & 0.8 & 0.0 & 0.6 \\
DTD & 0.1 & 0.0 & 0.1 & 0.7 & 0.7 & 0.7 & 0.7 & 0.8 & 0.8 & 0.8 & 0.8 & 0.8 & 0.9 & 0.9 & 0.9 & 0.0 & 0.6 \\
EMNIST & 0.1 & 0.0 & 0.1 & 0.8 & 0.7 & 0.7 & 0.7 & 0.7 & 0.8 & 0.8 & 0.8 & 0.8 & 0.8 & 0.9 & 1.0 & 0.1 & 0.6 \\
EuroSAT & 0.1 & 0.0 & 0.1 & 0.7 & 0.6 & 0.6 & 0.6 & 0.6 & 0.6 & 0.6 & 0.6 & 0.6 & 0.6 & 0.6 & 0.6 & 0.0 & 0.5 \\
FashionMNIST & 0.1 & 0.0 & 0.1 & 0.7 & 0.7 & 0.6 & 0.6 & 0.6 & 0.7 & 0.7 & 0.6 & 0.7 & 0.7 & 0.7 & 0.5 & 0.0 & 0.5 \\
FER2013 & 0.1 & 0.0 & 0.1 & 0.7 & 0.6 & 0.7 & 0.7 & 0.7 & 0.7 & 0.8 & 0.8 & 0.8 & 0.9 & 0.9 & 0.8 & 0.0 & 0.6 \\
Flowers102 & 0.1 & 0.0 & 0.1 & 0.5 & 0.5 & 0.4 & 0.5 & 0.5 & 0.4 & 0.4 & 0.4 & 0.5 & 0.5 & 0.5 & 0.5 & 0.0 & 0.4 \\
Food101 & 0.1 & 0.0 & 0.1 & 0.7 & 0.7 & 0.7 & 0.7 & 0.7 & 0.7 & 0.7 & 0.7 & 0.7 & 0.8 & 0.8 & 0.7 & 0.0 & 0.6 \\
GTSRB & 0.1 & 0.0 & 0.1 & 0.5 & 0.5 & 0.5 & 0.5 & 0.5 & 0.5 & 0.5 & 0.5 & 0.5 & 0.6 & 0.7 & 0.8 & 0.0 & 0.4 \\
KMNIST & 0.1 & 0.0 & 0.1 & 0.7 & 0.6 & 0.6 & 0.6 & 0.7 & 0.7 & 0.8 & 0.7 & 0.7 & 0.8 & 0.8 & 0.6 & 0.0 & 0.5 \\
MNIST & 0.1 & 0.0 & 0.1 & 0.6 & 0.6 & 0.6 & 0.6 & 0.6 & 0.6 & 0.6 & 0.6 & 0.6 & 0.6 & 0.7 & 0.9 & 0.0 & 0.5 \\
OxfordIIITPet & 0.1 & 0.0 & 0.1 & 0.6 & 0.6 & 0.6 & 0.6 & 0.6 & 0.6 & 0.5 & 0.5 & 0.5 & 0.6 & 0.6 & 0.5 & 0.0 & 0.4 \\
PCAM & 0.1 & 0.0 & 0.2 & 0.7 & 0.7 & 0.7 & 0.7 & 0.7 & 0.7 & 0.6 & 0.6 & 0.6 & 0.6 & 0.6 & 0.5 & 0.0 & 0.5 \\
RenderedSST2 & 0.1 & 0.0 & 0.2 & 0.6 & 0.6 & 0.6 & 0.6 & 0.6 & 0.6 & 0.6 & 0.6 & 0.8 & 0.9 & 0.9 & 0.8 & 0.0 & 0.5 \\
RESISC45 & 0.1 & 0.0 & 0.1 & 0.7 & 0.7 & 0.6 & 0.7 & 0.7 & 0.7 & 0.7 & 0.7 & 0.7 & 0.8 & 0.8 & 0.8 & 0.0 & 0.5 \\
STL10 & 0.1 & 0.0 & 0.1 & 0.6 & 0.6 & 0.5 & 0.5 & 0.5 & 0.5 & 0.5 & 0.5 & 0.5 & 0.5 & 0.5 & 0.4 & 0.0 & 0.4 \\
SUN397 & 0.1 & 0.0 & 0.1 & 0.7 & 0.7 & 0.7 & 0.7 & 0.7 & 0.7 & 0.8 & 0.8 & 0.8 & 0.9 & 1.0 & 1.0 & 0.0 & 0.6 \\
SVHN & 0.1 & 0.0 & 0.1 & 0.7 & 0.7 & 0.7 & 0.7 & 0.7 & 0.7 & 0.7 & 0.7 & 0.7 & 0.7 & 0.8 & 0.9 & 0.0 & 0.6 \\
\midrule
\multicolumn{15}{l}{\em DUNE benchmark} \\
NYU-d & 7.0 & 0.1 & 0.3 & 1.9 & 1.8 & 1.6 & 2.5 & 2.8 & 3.1 & 3.2 & 3.3 & 3.4 & 3.3 & 3.2 & 3.3 & 0.0 & 2.5 \\
ADE20k & 1.4 & 0.0 & 0.6 & 3.5 & 3.4 & 3.6 & 4.6 & 4.5 & 4.7 & 5.0 & 5.4 & 5.8 & 6.3 & 6.8 & 6.6 & 0.1 & 3.9 \\
MapFree & 6.9 & 0.2 & 2.4 & 22.7 & 26.7 & 23.1 & 23.0 & 22.6 & 22.8 & 23.4 & 23.9 & 24.5 & 25.1 & 25.4 & 25.7 & 0.0 & 18.6 \\
BEDLAM & 27.9 & 0.6 & 5.5 & 47.1 & 50.4 & 52.2 & 51.9 & 54.7 & 57.2 & 58.7 & 60.5 & 61.4 & 63.2 & 66.5 & 69.5 & 0.0 & 45.5 \\
\bottomrule
\end{tabular}
}
\end{table}